\documentclass[a4paper,fleqn]{cas-sc}

\usepackage[authoryear,longnamesfirst]{natbib}
\usepackage{colortbl}
\usepackage{multirow}
\usepackage{float}

\def\tsc#1{\csdef{#1}{\textsc{\lowercase{#1}}\xspace}}
\tsc{WGM}
\tsc{QE}
\tsc{EP}
\tsc{PMS}
\tsc{BEC}
\tsc{DE}

\begin{document}
\let\WriteBookmarks\relax
\def\floatpagepagefraction{1}
\def\textpagefraction{.001}

\shorttitle{MALM: A Multi-Information Adapter for Large Language Models to Mitigate Hallucination}

\shortauthors{Ao Jia et~al.}

\title [mode = title]{MALM: A Multi-Information Adapter for Large Language Models to Mitigate Hallucination}                      



%
\author[1]{Ao Jia}
[type=editor,
                        orcid=0009-0008-5405-3331
                        ]



\ead{aojia@bit.edu.cn}

 

\affiliation[1]{organization={Beijing Institute of Technology},
    city={Beijing},
    country={China}}

\author[1]{Haiming Wu}
\ead{wuhm@bit.edu.cn}

\author[1]{Guohui Yao}
\ead{ghyao@bit.edu.cn}



\author[1, 2]{Dawei Song}
\ead{dawei.song2010@gmail.com}
\cormark[1]

\cortext[cor1]{Corresponding author}

\affiliation[2]{organization={The Open University},
    city={Milton Keynes},
    country={United Kingdom}}
\author%
[3]
{Songkun Ji}
\ead{jisongkun@bistu.edu.cn}
\cormark[1]

\affiliation[3]{organization={Beijing Information Science and Technology University},
    city={Beijing},
    country={China}}

\author%
[4]
{Yazhou Zhang}
\ead{yazzhang@polyu.edu.hk}

\affiliation[4]{organization={The Hong Kong Polytechnic University},
    city={Hongkong},
    country={China}}



\begin{abstract}
Large language models (LLMs) are prone to three types of hallucination: \textit{Input-Conflicting}, \textit{Context-Conflicting} and \textit{Fact-Conflicting} hallucinations. The purpose of this study is to mitigate the different types of hallucination by exploiting the interdependence between them.
For this purpose, we propose a \textbf{M}ulti-Information \textbf{A}dapter for Large \textbf{L}anguage \textbf{M}odels (MALM). This framework employs a tailored multi-graph learning approach designed to elucidate the interconnections between original inputs, contextual information, and external factual knowledge, thereby alleviating the three categories of hallucination within a cohesive framework.
Experiments were carried out on four benchmarking datasets: HaluEval (35,000 samples), TruthfulQA (790 samples), Natural Questions (62,490 samples), and TriviaQA (71,726 samples).
We evaluated the proposed framework in two aspects: (1) adaptability to different base LLMs on HaluEval and TruthfulQA, to confirm if MALM is effective when applied on 7 typical LLMs. MALM showed significant improvements over LLaMA-2, e.g. 5.10\% (in HaluEval) and 132.01\% (in TruthfulQA) in terms of ROUGE-2; (2) generalizability to retrieval-augmented generation (RAG) by combining MALM with three representative retrievers (BM25, Spider and DPR) separately. The results demonstrated significant improvements over the state-of-the-art RAG models on the HaluEval, NQ and TriviaQA datasets, ranging from 17.57\% to 37.49\% in terms of ROUGE-1 and from 11.28\% to 27.92\% in terms of FEQA. Furthermore, automated and human evaluations were conducted to substantiate the correctness of experimental results, where GPT-4 and 3 human volunteers judged which response was better between LLaMA-2 and MALM. The results showed that both GPT-4 and human preferred MALM in 79.4\% and 65.6\% of cases respectively. The results validate that incorporating  the complex interactions between the three types of hallucination through a multilayered graph attention network into the LLM generation process is effective to mitigate the them. The adapter design of the proposed approach is also proven flexible and robust across different base LLMs.
\end{abstract}


\maketitle

\section{Introduction}
\subsection{Background}
In recent years, the rapid development of large language models (LLMs) has manifested their immense potential across various natural language processing (NLP) tasks, such as classification~\citep{maazallahi2025advancing}, retrieval~\citep{wang2024searching}, and generation~\citep{chang2024survey}. The tremendous performance of LLMs has garnered a widespread attention in the computing and information science community. However, during the generation process, LLMs occasionally produce information that appears correct but is, in fact, erroneous. This phenomenon is commonly referred to as the hallucination problem~\citep{ji2023survey}, which is recognized by more and more researchers~\citep{liu2025evopath}. As the popularity and overall performance of LLMs continue to rise, their outputs containing hallucinated information are becoming increasingly fluent and convincingly presented, which poses a challenge for humans to accurately discern such inaccuracies. This issue has serious implications for the effective deployment of LLMs in specialized domains such as healthcare and finance. Hence, the mitigation of hallucination has become a crucial research problem.

 \begin{figure}[b]
    \centering
  \includegraphics[width=0.7\textwidth]{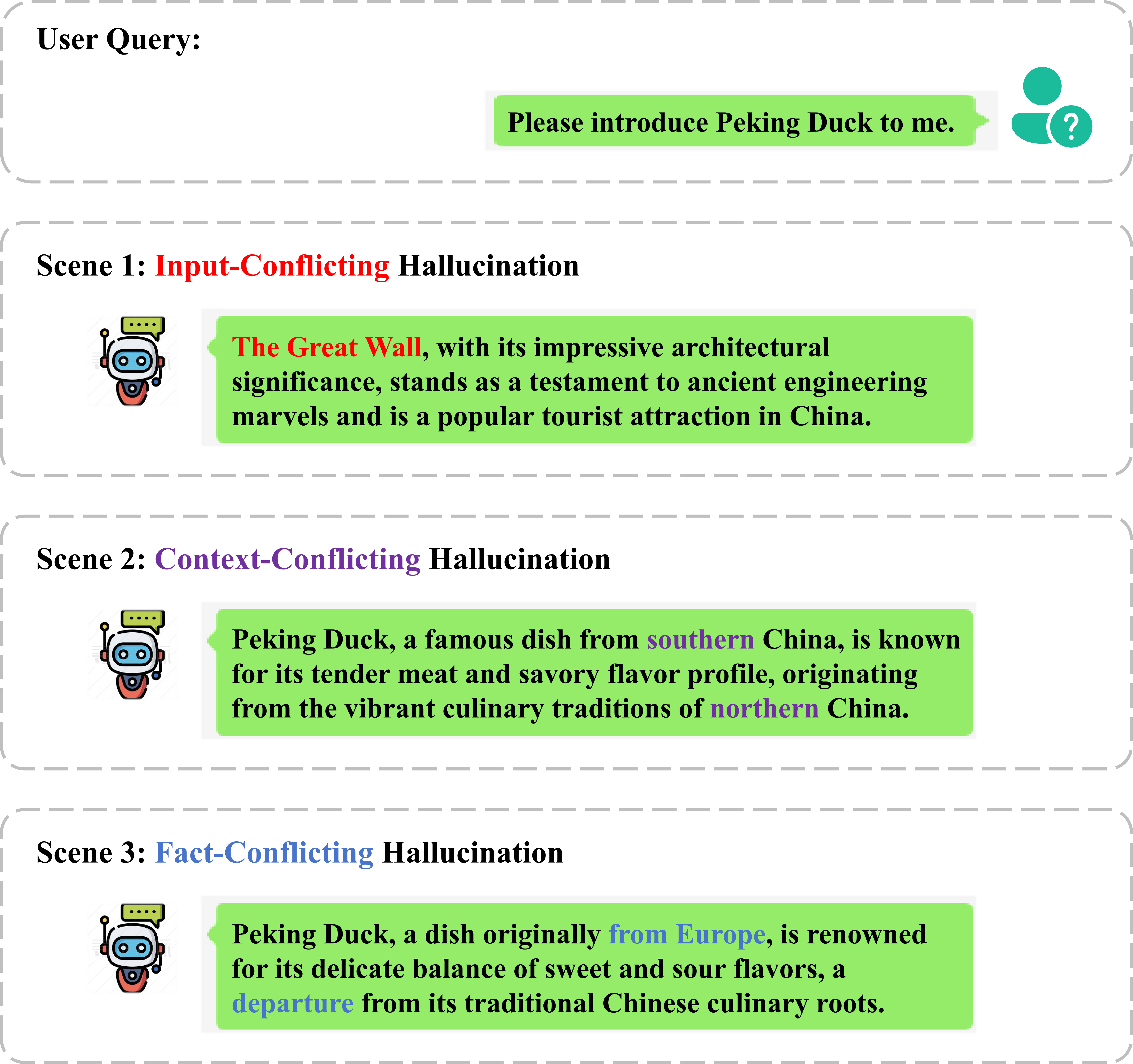}
  \caption{Examples of three types of hallucination.}
   \label{1-1}
 \end{figure} 

Hallucination can be categorized into three different types: Input-Conflicting hallucination, Context-Conflicting hallucination, and Fact-Conflicting hallucination~\citep{zhang2023siren}, as shown in Fig.~\ref{1-1}. Input-Conflicting hallucination refers to conflict between the LLM's response and the user's original input. This is illustrated in Scene 1, where the LLM should respond with content related to Peking Duck, rather than an introduction of the Great Wall. Scene 2 shows an example of Context-Conflicting hallucination, which refers to the inconsistencies within the content generated by the LLM. In the beginning of Scene 2, the response first states that Peking Duck derives from southern China, while it changes to northern China at the end, which is obviously self-contradictory. Finally, Fact-Conflicting hallucination means that the LLM's generated content contradicts with factual knowledge. For example, Scene 3 contains a palpable factual error that Peking Duck originates from Europe, while the correct answer should be China. On the other hand, it still mentions Peking Duck as a traditional Chinese delicacy, thus not straying too far from the factual context.

Recent studies have made progress to detect and mitigate different types of hallucination, in explicit or implicit ways. For instance, \cite{lee2022factuality} designed a dataset and the corresponding metrics to evaluate the factual accuracy of LLMs. Taking into account contextual information, a TopicPrefix was introduced to address erroneous associations between entities that arise from the fragmentation of document chunks during LLM training. \cite{manakul2023selfcheckgpt} proposed a method, called SelfCheckGPT, to detect hallucination by multiple sampling of responses.
This approach validates the consistency of the generated responses to ascertain whether they were self-contradictory, leveraging contextual information to a certain extent. 
\cite{leiser2024hill} proposed HILL to support users in identifying hallucinations in LLMs.
\cite{mcdonald2024reducing} reduced the hallucination with the application of knowledge distillation, using temperature scaling and intermediate layer matching.
Moreover, various RAG (Retrieval-augmented Generation)~\citep{chen2024benchmarking} approaches have been shown promising, which combine information retrieval  generation techniques with LLMs and makes use of the retrieved external knowledge to enable LLMs to generate answers with reduced factual hallucination. \cite{ramrakhiyani2025gauging} also emphasized the importance of knowledge to LLMs.
However, these existing methods neglected the complex interactions between the three types of hallucination, resulting in the inability to comprehensively mitigate hallucination from multiple angles simultaneously. The accuracy of LLM-generated responses is jointly influenced by these three interconnected angles, such that the improvement in one can facilitate the model's ability to better manage the other two. For instance, in Fig.~\ref{1-1} Scene 2, if the LLM has sufficient knowledge that Peking Duck is from northern China, then the self-contradictory phenomenon would not appear. In addition, reducing Input-Conflicting hallucination can help awaken the LLM's knowledge about the input information more precisely. For example, in Scene 1, if the LLM understands that the user is asking about Peking Duck rather than the Great Wall, it will likely provide a description that better matches the user's intention.

\subsection{Research Objectives}
The main objective of this research is to mitigate the Input-Conflicting, Context-Conflicting and Fact-Conflicting hallucination problems in LLM generation. To achieve this goal, the study aims to propose a method to simultaneously consider the interdependence of the three types of hallucination within a unified framework. 

The secondary objective is to integrate the proposed framework with retrieval-augmented generation (RAG). The documents retrieved by a retriever can serve as external knowledge in the proposed framework to enhance the factual accuracy of generation. As an individual method from RAG, MALM can be used to complement existing RAG systems and can improve their performance using the originally retrieved documents without any extra data augmentation. Hence, MALM also decreases the influence of the over-reliance issue on data in RAG.

\subsection{Proposed Method}
In this paper, we take the first step to overcome the aforementioned limitation by leveraging the correlations between the input, context and factual knowledge for hallucination mitigation, instead of treating them as three individual problems. We thus propose MALM, a \textbf{M}ulti-Information \textbf{A}dapter for Large \textbf{L}anguage \textbf{M}odels, to jointly mitigate the three types of hallucination by using graph neural networks. Specifically, the adapter consists of multi-layered graphs. Each layer contains three subgraphs respectively corresponding to the user's input, the LLM's response and the external knowledge. In each subgraph, the nodes represent tokens, while the edges denote connections among these tokens. To facilitate the complex interactions among different types of information, three types of connections are established: input connections, context connections, and knowledge connections. Specifically, the input connections involve uni-directional edges from the input to output nodes, aiming to narrow the gap between the input query and the output response. The context connections link the output nodes with each other via a masked fully connected graph. The knowledge connections aim to provide references to external knowledge, where each output node is connected with all knowledge nodes. These connections enable the propagation and aggregation of relevant information across different types of nodes. The adapter can be slotted between the transformer blocks and the output head of LLMs. Thereby effectively combining the language processing and information extraction capabilities of LLMs with the multi-type information interaction capabilities of graph neural networks. Moreover, the design of adapter allows that MALM can be widely used in various applications and different LLMs for hallucination mitigation. Because the main design purpose of MALM is the flexibility for different LLMs as a plug-in adapter, which makes MALM to be a universal method for almost every applications where LLMs can be applied. And it also can be used as an extra step in RAG methods, to better help LLMs leverage the retrieved knowledge information.

Two sets of experiments are carried out to evaluate our proposed method. The first set is to test the effectiveness and adaptability of MALM over different LLMs on two benchmark datasets, i.e., HaluEval~\citep{li2023halueval} and TruthfulQA~\citep{lin2022truthfulqa}. It is important to note that HaluEval is designed specifically for evaluating the performance of LLMs in recognizing hallucination and directly provides the relevant external knowledge for each input question. The experimental results indicate that MALM yields significant improvements compared to a wide range of typical LLM baselines. The second set of experiments is to explore the effectiveness of MALM in RAG, where certain retrievers are used to obtain external knowledge. Given that the automatically retrieved external documents may not always be relevant to the input query, testing MALM in this scenario reflects its robustness and generalizability to RAG. This set of experiments are conducted on three datasets, i.e., HaluEval, NQ~\citep{kwiatkowski-etal-2019-natural} and TriviaQA~\citep{joshi2017triviaqa}, and the results demonstrate that MALM outperforms the state-of-the-art RAG models. Additionally, we perform an ablation study to show the contributions of different components of MALM to the overall performance and conduct an experiment of models layers to observe the effects of layers. Furthermore, automated evaluation with GPT-4 and human evaluation are carried out. The results further validate the superiority of MALM. 

The rest of this paper is organized as follows. In Section 2, we comprehensively summarize the related work. Section 3 describes the proposed MALM method in detail. In Section 4, we report the empirical experiments and analyze the results. Section 5 concludes the paper and highlights future research directions.

\section{Related Work}\label{two}
\subsection{Hallucination Detection and Mitigation}
The most intuitive method for hallucination detection and mitigation is the so-called verification and revision. \cite{dhuliawala2024chain} developed a method named Chain-of-Verification, which first verifies the generated response and then regenerates a revised response based on the verification result. Similarly, \cite{varshney2023stitch} proposed to validate the uncertain concepts in the response by retrieving relevant external knowledge. \cite{mundler2024self} recognized the importance of context information. They employed prompts containing contextual description to guide LLMs to trigger, detect and mitigate self-contradictions of the following statements based on the description. Another method, namely RARR~\citep{gao2023rarr}, devised a research stage and a revision stage to automatically fix the unsupported content.

While the aforementioned methods can correct the hallucination to some extent, they did not enhance the capability of LLMs so as to avoid hallucination. In contrast, \cite{chen2023purr} trained a model namely PURR, by denoising a corrupted statement. It could be used to edit the ungrounded statement with the help of relevant evidence. Furthermore, for generative models, \cite{sun2023contrastive} employed contrastive learning, where the negative and positive knowledge were mixed up in the training. This method reduced the generation probability of the confusing negative knowledge in LLMs, thereby eliminating the hallucination.

It is worth noting that Multimodal Large Language Models (MLLMs) are also plagued by the critical issue of hallucination, and there has been a series of recent work dedicated to solving this problem. HACL~\citep{jiang2024hallucination} used hallucination augmented contrastive learning to increase the distance between visual representations and hallucinative text in the representation space. This method favors MLLM on hallucination mitigation and visual comprehension. UNIHD~\citep{chen2024unified} aimed to identify and detect modality-conflicting and fact-conflicting hallucination in both image-to-text and text-to-image generation. \cite{huang2024opera} found that MLLMs tend to generate hallucinative tokens by over-trusting the closer summary tokens and ignore most of the previous tokens and vision tokens. To address the problem, a method namely OPERA, was proposed by introducing an over-trust logit penalty and retrospection-allocation strategy.

However, none of these methods take into account all three types of hallucination. Moreover, the time cost of the two-step approach (i.e., verification and revision) is high. Instead, mitigating hallucination during generation can be more efficient. In this paper, we propose a multi-layer graph based adapter, i.e. MALM, to fully capture the interaction between three types of hallucination, and employ it on the generative models to address the hallucination problem in a one-step way.

\subsection{Graph Neural Networks}
Graph Neural Networks (GNNs) represent data into the form of graph, where each node is associated to its neighbors through connections of multiple types~\citep{wu2020comprehensive}. GNNs are widely used in the information science research because information can be propagated and aggregated interactively through the graph structure~\citep{zhou2020graph, hao2024simplices}. Since the emergence of GNN~\citep{gori2005new}, various genres of GNN are designed, such as GCN (Graph Convolutional Networks)~\citep{kipf2017semisupervised}, GraphSAGE~\citep{hamilton2017inductive}, and GAT (Graph Attention Networks)~\citep{velickovic2018graph}. These works have shown a convincing performance in many tasks, such as link prediction~\citep{wang2024efficient}, summarization~\citep{chen2024entity}, sentiment analysis~\citep{yin2024textgt}, and graph data augmentation~\citep{wu2025asymmetric}. The characteristics of the graph structure enable a better interaction and fusion of information. For instance, \cite{xing2022darer} built a dual-task reasoning temporal graph (DRTG) for integrating information between the tasks of sentiment and act prediction. \cite{zhang2023m3gat} constructed an interactive conversation graph layer to learn multi-modal complementarity and knowledge across tasks. \cite{meng2022gnn} applied graph neural network on the top of a vanilla LLM to facilitate the interactions between the context and retrieved reference. HGNRec~\citep{li2024homogeneous} split a heterogeneous graph containing two different types of nodes (i.e., Apps and third-party libraries) into two homogeneous graph neural networks, one for each type. This method facilitated the propagation and aggregation of information within the same knowledge space.

It has been demonstrated that GNNs can serve as effective adapters for LLMs~\citep{huang2024can}. In our work, we adopt GAT as the architecture of the adapter module for its brilliant ability to quantify the relationships between different types of information. We take the first step to holistically mitigate the three types of hallucination problem by constructing input connections, context connections and knowledge connections into a unified graph framework.

\subsection{Retrieval-Augmented Generation}
Retrieval-augmented generation (RAG)~\citep{izacard2021leveraging,ram-etal-2023-context,shi-etal-2024-replug} combines the capabilities of text generation and information retrieval, allowing LLMs to retrieve relevant information from external data sources to alleviate their shortcomings in memory and external knowledge. This approach has potential advantages in improving the accuracy, explanability, and real-time performance of LLM-based text generation, and has therefore attracted widespread attention. To date, the majority of research related to RAG has focused on the processes of input~\citep{zhou2023least}, retrieval~\citep{sarthi2024raptor}, and generation~\citep{cheng2024lift}. For example, to address the issue of insufficient information caused by short and discontinuous retrieval documents, RAPTOR~\citep{sarthi2024raptor} proposed a solution that processed the entire retrieval dataset using clustering and text summarization techniques to construct a hierarchical retrieval dataset. SELF-RAG~\citep{asai2024selfrag} proposed an on-demand retrieval strategy that trains large language models (LLMs) to independently determine whether retrieval is necessary, thereby improving the quality of the generated content. RECOMP~\citep{xu2023recomp} suggested that when retrieved documents are too long, it can affect the model's efficiency and the quality of the generated content. To address this problem, the model compressed the retrieved documents into text summaries. On the other hand, since RAG's performance heavily depends on the retrieved documents, CRAG~\citep{yan2024corrective} used an error correction model to assess the relevance of the retrieved documents to the query, and subsequently processed the documents in three hierarchical levels. Although these methods have somewhat improved the model's performance, they overlook the connections and interactions between the query input, the context of response generation and the external knowledge, resulting in sub-optimal performance.

Graph RAG~\citep{edge2024local} was an RAG method recently proposed by Microsoft. It built a structural and hierarchical knowledge graph to enhance the LLMs' output. This differed from the traditional RAG methods that typically used vector-based similarity computation in the retrieval process. 
Methodologically Graph RAG is consistent with our approach, but differs in the aim and how the graph structure is used. The aim of Graph RAG is to generate answers towards questions, where the graph structure is used to build the relationship of knowledge, and hallucination is mitigated implicitly. Moreover, it does not explicitly take into account different types of hallucination and their complex interactions. Our approach uses the graph structure to explicitly capture the intra- and inter-relationships between different types of information underlying multiple types of hallucination, thereby effectively mitigating them in a unified framework.

\subsection{Modular Language Models and Adapter}
Large language models can achieve outstanding performance on a wide range of NLP tasks through fine-tuning. However, in specific fields, a large amount of high-quality data and sufficient computing power are required to fine-tune a sufficiently useful model, which is relatively difficult. To address this challenge, various modules are introduced to extend the existing pretrained LLMs. Compared with fine-tuning the whole language model, training smaller modules can be a more parameter-efficient way to obtain a task-specific model. Adapter is one of the most commonly used modules adding into the LLMs. It outperforms fine-tuning on low-resource tasks, and is less prone to overfitting~\citep{he2021effectiveness}. \cite{houlsby2019parameter} designed an adapter with a bottleneck architecture, and integrated it into each Transformer layer. This research proved that adapter-tuning can achieve remarkable results that are comparable to full parameter fine-tuning by using fewer parameters. \cite{pfeiffer2023mmt5} identified a "source language hallucination problem", i.e., the model generated text with true meaning but the language was different from the input, which can be categorized into the Input-Conflicting Hallucination. To tackle the problem, a mmT5 was proposed, in which language-specific moduleswere used to improve the multilingual capacity of T5. The work also proved the effectiveness of using adapters in LLMs.

In this paper, MALM is designed as an adapter to address the hallucination problem in LLMs. It has the advantages in multiple ways: (1) Low Computing Cost. We can expand the capability of the LLMs by fine-tuning MALM with fewer parameters; (2) High Training Efficiency. Better results can be achieved in less training time, so we also have more opportunities to adjust the structure of the entire framework; (3) High Adaptability and Generalizability. By the design of the lightweight plug-in structure, MALM can be widely applied in various LLMs. The adapter nature also allows a flexible generalization of MALM to different RAG methods.

\section{The Proposed Approach}

\begin{figure*}[ht]
\centering
\includegraphics[width=1.0\textwidth]{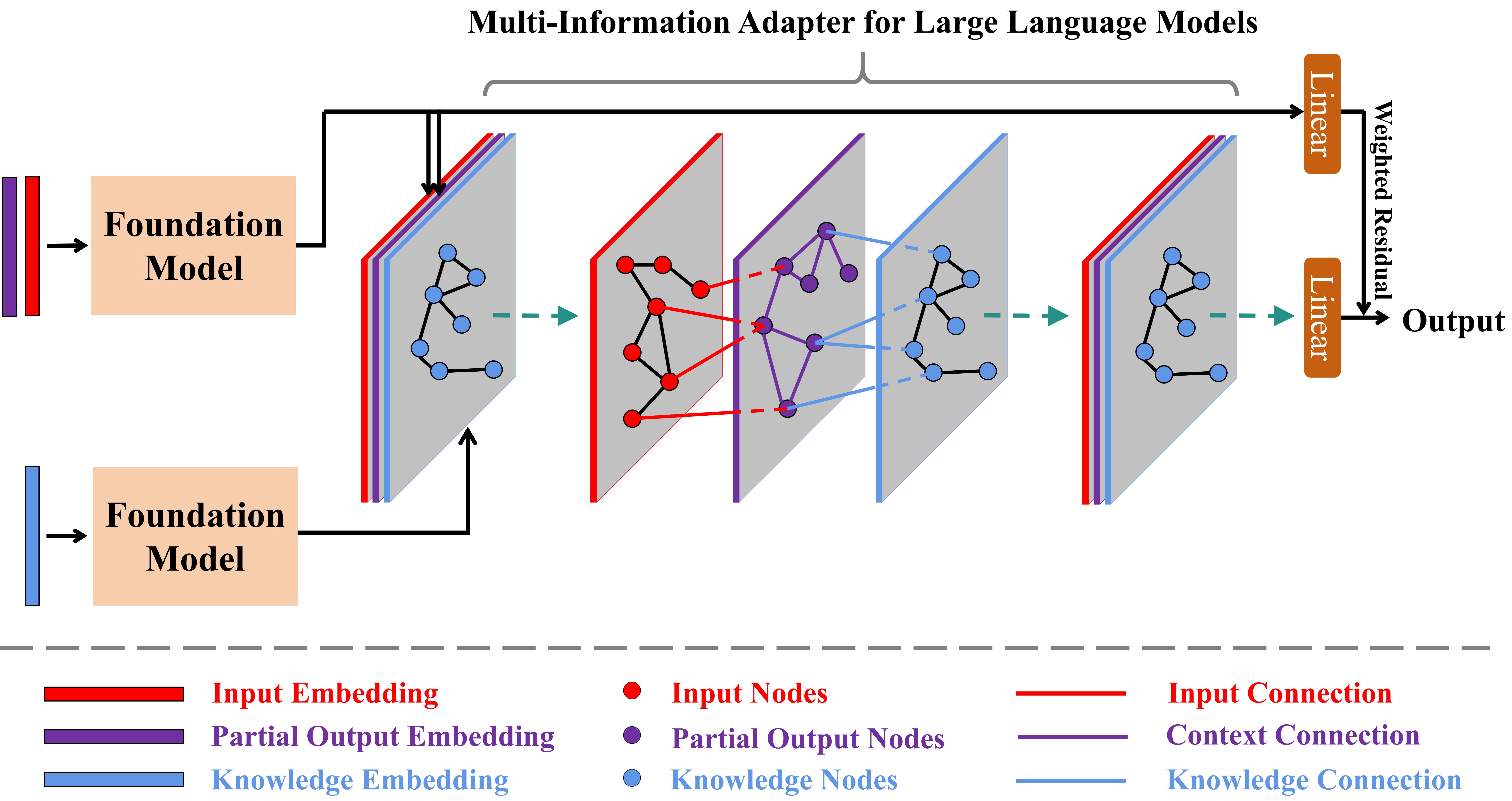}
\caption{The overall architecture of the proposed approach.} \label{model}
\end{figure*}

As discussed in the previous section, recent models have attempted to tackle the hallucination problem so that the LLM-generated responses can be faithful to the user intent, context and factual knowledge as much as possible. However, the existing approaches have not considered the aforementioned three types of information and their relationships in a unified framework. In this paper, we propose MALM, which is an adapter aiming to leverage multiple types of information in a variety of foundation models. Hence, it is designed as a plug-in GAT (Graph Attention Networks) structure, which can conveniently slot into different LLMs.

Fig.~\ref{model} depicts the overall architecture of our proposed approach, which consists of two main parts: foundation model and the multilayer MALM. MALM is inserted between the last transformer block of the foundation model.

To begin with, the embeddings of different types of information (i.e. input embedding, partial output embedding, and knowledge embedding) are fed into the foundation model. And we can obtain the hidden states of each token from the last transformer block in the foundation model. And the hidden states are used to initialize the features of the vertices in the first graph layer of MALM. Each graph layer contains three types of subgraphs, corresponding to the three types of information we are concerned about (i.e., the user's input, the context in the LLM's output and the factual knowledge). In each subgraph, the nodes and edges respectively represent the tokens and the connections between them (for the sake of simplicity and aesthetics, we don't display all the edges between nodes in Fig.~\ref{model}). Homogeneous information can be synthesized in the same subgraph. The connections between different subgraphs enable sufficient interaction of heterogeneous information. After updating in multiple layers, we input the updated hidden states from the MALM into the LLM’s linear layer to get the prediction. And we also use the original hidden states from the last block of the LLM to get the original prediction. Then a weighted residual connection mechanism is used to balance the prediction from MALM and the original prediction. Subsequently, we can get the final prediction for the next token.
In the rest of this section, we will provide a detailed description of the proposed approach.

\subsection{Problem Formulation}
Let us assume that there is a sample $E$ represented as $E = \left \{ T, K, Y \right \}$, where $T$ denotes the user's input query or question, $K$ denotes the factual knowledge relevant to the input, and $Y = \left \{ y_{1}, y_{2},...,y_{n} \right \}$ denotes the ground truth. $T = \left \{ t_{1}, t_{2},...,t_{m},...,t_{M} \right \}$, where $t_{m}$ represents the $m^{th}$ token of the input and $M$ represents the total number of tokens (i.e., words) in $T$. $K = \left \{ k_{1}, k_{2},...,k_{s},...,k_{S} \right \}$, where $k_{s}$ represents the $s^{th}$ token of the factual knowledge and $S$ represents the total number of tokens in $K$.

In general, LLM generates response $X = \left \{ x_{1}, x_{2},...,x_{n} \right \}$ when receiving $T$ as the input. $X_{<i} = \left \{ x_{1}, x_{2},...,x_{i-1} \right \}$ is the partial output sequence during the generation process, which is also the context information for the prediction of the next token.
Then, the research problem is summarized as: \emph{Given an input $T$, a partial output $X_{<i}$ as context and factual knowledge $K$, how to leverage the inter-relatedness across different types of information to mitigate hallucination?} It can be formulated as:

\begin{equation}
\zeta = \prod_{i}p\left ( x_{i}\mid T, K, X_{<i}, \mathcal{G}, \Theta \right )
\end{equation}
where $\mathcal{G}$ represents the correlation between the input information $T$, context information $X_{<i}$ and factual knowledge $K$, and $\Theta$ denotes the parameter set.

The main notations and their descriptions are listed in Table~\ref{tab:notation}.

\begin{table*}[ht]
\begin{center}
\caption{Main notations and their descriptions.}
\label{tab:notation}
\scalebox{1.0}{
\begin{tabular}{c|p{0.35\textwidth}}
\hline
Notation            & Description                                             \\ \hline
$T$                   & Input of the foundation model, i.e., the user query.     \\
$t_m$                  & $t_m$ is the $m^{th}$ token of $T$.                               \\
$T^0, t^0_m$             & The initial embedding of $T$ and $t_m$ in the graph layer.                       \\ \hline
$K$                   & The factual knowledge related to the query.              \\
$k_s$                  & $k_s$ is the $s^{th}$ token of $K$.                               \\
$K^0, k^0_s$             & The initial embedding of $K$ and $k_s$ in the graph layer.                       \\ \hline
$X$                   & The generated response towards the query.                \\
$x_i$                  & $x_i$ is the $i^{th}$ token of $X$.                               \\
$X_{<i}$       & Partial output sequence during the generation process. $X_{<i} = \left \{ x_{1}, x_{2},...,x_{i-1} \right \}$   \\
$X^0_{<i}, x^0_i$ & The initial embedding of $X_{<i}$ and $x_i$ in the graph layer.           \\ \hline
$y_i$                  & The ground truth.                                        \\
$\hat{y}_i$               & The prediction of the model.                             \\ \hline
$A$                   & The adjacent matrix of the graph.                        \\
$A_{pq}$                 & When node p connects to node q, $A_{pq}=1$. Otherwise, $A_{pq}=0$. \\ \hline
\end{tabular}
}
\end{center}
\end{table*}

\subsection{Foundation Model}
We design MALM as a plug-in module and the purpose is to flexibly apply it to different LLMs. In other words, LLMs are the foundation of MALM. Hence we refer the selected LLM to as the foundation model.

As shown in Fig.~\ref{model}, we duplicate the foundation model into two parallel streams. One is used to obtain features for the input $T$ and the partial output $X_{<i}$. The other is used to obtain features for the factual knowledge $K$. Specifically, $T^{0} = \left \{t_{1}^{0}, t_{2}^{0}, \dots, t_{M}^{0} \right \}$, $X_{<i}^{0} = \left \{ x_{1}^{0}, x_{2}^{0}, \dots, x_{i-1}^{0} \right \}$ and $K^{0} = \left \{ k_{1}^{0}, k_{2}^{0},...,k_{S}^{0} \right \}$, where $T^{0} \in \mathbb{R} ^{M \times d}$, $X_{<i}^{0} \in \mathbb{R} ^{(i-1) \times d}$, $K^{0} \in \mathbb{R} ^{S \times d}$. The superscript $d$ denotes the dimensionality of vertex representation and $0$ means they are the initial sequence of vertex representation for the MALM. Because the input and partial output may have already contained hallucinated information, using a separate parallel  stream of the foundation LLM to obtain the features of the factual knowledge can alleviate the issue.

\subsection{Multi-Information Adapter for Large Language Models}
The proposed Multi-Information Adapter for Large Language Models, namely MALM, consists of multiple layers of graph structure. In each layer, we build input connections, context connections and knowledge connections to facilitate the interactions among different types of information.

We present the following notation of a directed graph $\mathcal{G}$ with vertices and edges, which can be formalized as:
\begin{equation}
\mathcal{G} = \left ( \mathcal{V}, \mathcal{E}, \mathcal{W}, \mathcal{A} \right )
\end{equation}
where the vertices $\upsilon_{j}, \upsilon_{k} \in \mathcal{V}$, the edge $\overrightarrow{\left \langle v_{j}, v_{k}\right \rangle} \in \mathcal{E}$, $\alpha_{jk} \in \mathcal{W}$ is the weight of edge $\overrightarrow{\left \langle v_{j}, v_{k}\right \rangle}$, s.t. $0 \le \alpha_{jk}\le 1$. $\mathcal{A}$ is the adjacent matrix of the graph, which is initialized as a zero matrix, and $\mathcal{A} \in \mathbb{R} ^{(M+i-1+S) \times (M+i-1+S)}$.

Each layer of the graph contains three subgraphs, representing the three types of information, i.e., input information, context (i.e., partial output) information and factual knowledge, respectively. The details of subgraph construction are depicted as follows.

\textbf{Vertices.~}
Each token in the input, partial output and factual knowledge is considered as a vertex in the corresponding subgraph, satisfying $T, X_{<i}, K \subseteq \mathcal{V}$. There are a total of $M+(i-1)+S$ vertices in the graphical structure, where $M$ vertices are built for input, $i-1$ vertices are built for partial output and $S$ vertices are built for factual knowledge. The features of vertices are initialized with $T^{0}, X_{<i}^{0}$ and $K^{0}$, where we have obtained from the two parallel streams of foundation model.

\textbf{Edges.~}
To capture the interactions among different types of information, three types of edges are constructed, including input, context and knowledge connections.

\textbf{Input Connections.~}
The input connections are established between the input subgraph and the partial output subgraph. Given that the model's outputs must align with the input query, it is essential to activate an interaction between input and output information to ensure consistency. Specifically, each node from the input subgraph is connected to the nodes from the partial output subgraph. In other words, an input connection is a directed link from input to output. It aims at preventing the output that may have been hallucinated from contaminating the input information in reverse. Formally, $\mathcal{A}_{p,q} = 1$ when $p \in T$ and $q \in X_{<i}$. The input connections facilitate the interaction between the output and the input, which makes the output more aligned with the user intent, thereby mitigating the input hallucination.

\textbf{Context Connections.~}
Context connections are constructed internally within the partial output subgraph, aiming at averting the phenomenon of self-contradiction within the model's output. For this purpose,  a masked fully connected graph is designed. The connection from a subsequent token to its preceding token is masked to prevent the former token from obtaining the future information, which violates the sequential nature of text generation. Formally, $\mathcal{A}_{p,q} = 1$ when $p, q \in X_{<i}$ and $p \geq q$.

\textbf{Knowledge Connections.~}
Knowledge connections introduce factual knowledge information during the text generation. Each node in the knowledge subgraph corresponds to a token in the description of the knowledge and is connected to all nodes within the partial output subgraph. A knowledge connection is also directed from the knowledge nodes to the partial output nodes, to avoid the factual hallucination.
This connection is important because hallucination most likely appears when the LLM lacks certain factual knowledge. Therefore, the introduction of knowledge connections can help reduce the fact-confliciting hallucination. Formally, $\mathcal{A}_{p,q} = 1$ when $p \in K$ and $q \in X_{<i}$.

Moreover, the vertices within the input subgraph are fully connected, and so are the vertices within the knowledge subgraph.

\textbf{GAT Features Updating.~}
To better utilize the three types of connections, we make some modifications to the updating process of GAT, which can be formulated as:
\begin{gather}
     q^{l}=\mathbf{W}^{l}\mathop{\|}\limits_{h=1}^H\sigma \left ( \sum_{p\in \mathcal{N}_{q}}\alpha _{pq}^{h,l}\mathbf{W}^{h,l}p^{l-1} \right )\label{eq:1}\\
     \mathcal{N}_{q}=\left \{ p \mid \mathcal{A}_{p,q} = 1 \right \}\label{eq:2}
\end{gather}
where $l\in \left \{ 1,2,3...,L \right \}$, $L$ is the number of graph layers. $q^{l} \in \mathbb{R} ^{d}$ represents the feature of a node in the $l$-th layer. $H$ is the number of heads for the attention mechanism. $||$ is concatenation operation. $\sigma$ denotes a nonlinear activation function. $\alpha _{pq}^{h,l}$ is the attention score of $p$ and $q$ in the $h$-th attention head of the $l$-th layer. $\mathbf{W}^{l}$ and $\mathbf{W}^{h,l}$ are the linear transformation's weights, and $\mathbf{W}^{l} \in \mathbb{R} ^{d \times (h \times d^{\prime})}, \mathbf{W}^{h,l} \in \mathbb{R} ^{d^{\prime} \times d}$. $\mathcal{N}_{q}$ denotes the set of neighbors of $q$.

The attention score $\alpha _{pq}^{h,l}$ can be calculated as Eq.~\ref{eq:attention1} and Eq.~\ref{eq:attention2}. $\vec{\mathbf{a}}$ is a column vector and $W$ is a matrix. Both of them are trainable. Through these parameters and algorithms, models can learn how to measure the importance from $p$ to $q$, and can automatically quantify the degree of influence of $p$ in terms of all the neighbors of $q$. When using the attention scores in Eq.~\ref{eq:1}, MALM can aggregate the non-hallucinated information to a node from all its neighbors in various types. Compared to traditional methods, MALM can simultaneously increase the proportion of input information, context information, and knowledge information, ensuring the reduction of hallucinated information. Hence, MALM can effectively use this mechanism to address how to utilize the three types of connection to mitigate the hallucination degree of nodes.
\begin{gather}
    \mathcal{F}\left ( \vec{i},\vec{j} \right )=LeakyReLU\left ( \vec{\mathbf{a}}^{T}\left [ \mathbf{W}\vec{i}|| \mathbf{W}\vec{j}\right ] \right )\label{eq:attention1}\\
    \alpha _{pq}^{h,l}=\frac{exp\left ( \mathcal{F}\left ( p^{l-1},q^{l-1} \right ) \right )}{\sum_{r\in \mathcal{N}_{k}}exp\left ( \mathcal{F}\left ( r^{l-1},q^{l-1} \right ) \right )}\label{eq:attention2}
\end{gather}

\begin{figure}[h]
\centering
\includegraphics[width=0.6\textwidth]{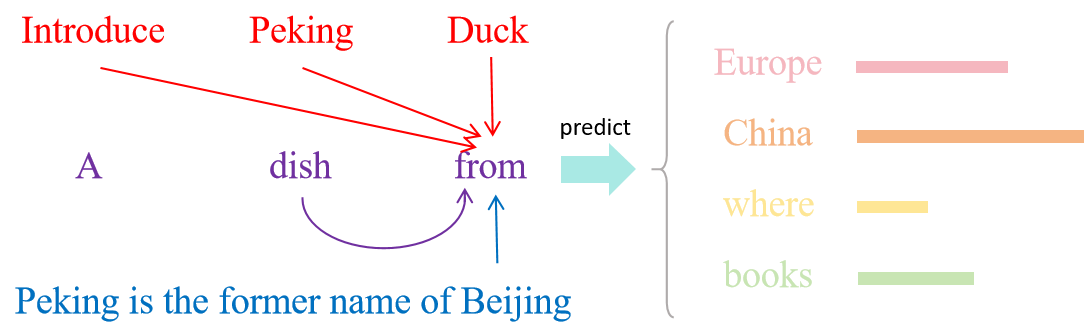}
\caption{An example to better explain the mechanism of MALM. The red, purple, and blue tokens represent input nodes, partial output nodes, and knowledge nodes in the graph, respectively. And the red, purple, and blue arrows represent input connection, partial output connection, and knowledge connection, respectively. Please note that not all edges are drawn for the sake of simplicity.} \label{exp}
\end{figure}

The updating process of features in GAT can promote the propagation and integration of information. Fig.~\ref{exp} is an example to better explain the mechanism of MALM. The red, purple, and blue tokens represent input nodes, partial output nodes, and knowledge nodes in the graph, respectively. And the red, purple, and blue arrows represent input connection, partial output connection, and knowledge connection, respectively. When updating the features, the input connection propagates the input information into the last output token "from", which guides model's prediction of the next token tending to align with the input intention, i.e., introducing. Thus there will be a smaller probability of generating interrogative words such as "where". The context connection form token "dish" delivers the context information to the token "from". The "from" token merges the information from "dish", so the model knows better that what needs to be introduced is a dish, not books or other things. And the knowledge tells that "Peking is the former name of Beijing". The token "Beijing" in the knowledge gives the information that the dish to be introduced is more related to China, rather than Europe. Overall, MALM uses the connection mechanism to mitigate the probability of generating hallucination, and successfully predicts the next token "China".

\subsection{Model Generation}
After $L$ layers updating, we can get the final representation of $x_{i-1}^{L}$. Then a linear layer is employed to get the prediction for the next token. Furthermore, we also use a weighted residual strategy to balance the graph prediction and original LLM prediction, which can be formulated as:
\begin{gather}
     y_i^{g}=Wx_{i-1}^{L}\label{eq:3}\\
     y_i^{o}=Wx_{i-1}^{0}\label{eq:4}\\
     \hat{y}_i=Softmax(\lambda y_i^{g}+(1-\lambda) y_i^{o})\label{eq:5}
\end{gather}
where $W$ is the linear transformation's weight matrix, $\lambda \in [0,1]$ is a hyperparameter.

\subsection{Model Training}
We use cross entropy as the loss function $\zeta$ for training:
\begin{equation}
    \zeta \triangleq -\sum_{i}\sum_{n}y_{i}^{n}\log{\hat y_{i}^{n}}
\end{equation}
where $y_{i}$ denotes the ground truth, $n$ represents the $n$-th dimension in a vector.

\section{Experiments}
In this section, we conduct two main experiments: (1) \textbf{Adaptability on Different LLMs.} MALM, as an adapter, can be applied to different foundation LLMs. Thus we first test the adaptability of MALM and verify whether it can improve performance for different foundation LLMs; (2) \textbf{Generalizability to RAG.} Since MALM utilizes the external knowledge, it can be naturally applied as an alternative RAG method. We combine MALM with different retrievers and use the retrieved documents as the knowledge information for MALM. We then test the effect of MALM on different retrievers and compare with the state-of-the-art RAG methods. In the following subsections, we first describe the experimental settings. Then, for each experiment, we introduce the datasets and baselines used. Finally we present and analyze the experimental results.

\subsection{Experimental Settings}

\subsubsection{Datasets}
We conduct experiments on four datasets, including HaluEval~\citep{li2023halueval}, TruthfulQA~\citep{lin2022truthfulqa}, Natural Questions (NQ)~\citep{kwiatkowski-etal-2019-natural} and TriviaQA~\citep{joshi2017triviaqa}.

HaluEval is specifically designed for research on hallucination. It not only includes hallucination annotations but also provides relevant external knowledge information for each input query. We can directly use the knowledge information included in the dataset to avoid the noise caused by the retrieval process, so that we can focus on the generation effectiveness. HaluEval contains 35,000 samples that are randomly selected from multiple datasets. The samples are categorized into four types, i.e., \textit{general}, \textit{question answering}, \textit{knowledge-grounded dialogue}, and \textit{text summarization}. In our work, we use \textit{question answering} to fine-tune and validate the models. There are 10,000 samples in \textit{question answering}, from which  9,500 samples are randomly selected for fine-tuning the models and the remaining 500 samples are used for validation. The questions are treated as user inputs, and the right answers are treated as ground truth responses.

Compared to HaluEval, which is specifically designed for the hallucination problem, TruthfulQA and NQ are designed for question answering (QA), and TriviaQA is designed for reading comprehension (RC). TruthfulQA is constructed for measuring how truthful models are, hence it can also be used for the research in hallucination. It consists of 790 samples, including questions, answers, and the URL sources of the answers. The questions are manually written by the authors aiming to elicit falsehood by humans and models, and are highly diverse with 38 categories. In our experiment, TruthfulQA is randomly split to 632 training samples and 158 test samples. Moreover, for the comprehensiveness of the experiment, we choose the rest two datasets, i.e., NQ and TriviaQA, to test the generalizability of our approach to RAG on different tasks . After pre-processing, the NQ dataset contains 58,880 training and 3,610 test samples. The questions are collected from the Google search engine. Heuristics are used to guarantee that the questions are natural and are from people who are sincerely looking for some kind of information. TriviaQA  consists of question-answer-evidence triples. The question-answer pairs are gathered from 14 trivia and quiz-league websites, and the evidence is obtained from Web search and Wikipedia articles. The dataset we use contains 60,413 samples for training and 11,313 for testing after pre-processing. For retrieval of external relevant documents that facilitate RAG, we use the English Wikipedia snapshot from December 20, 2018 as the source corpus, following~\cite{karpukhin2020dense}. Each article in the corpus is separated into text blocks of the length of 100 words, resulting in 21,015,324 passages.

\subsubsection{Evaluation Metrics}
We adopt the commonly used ROUGE-1, ROUGE-2, ROUGE-L~\citep{lin2004rouge}, Exact Match~\citep{rajpurkar2016squad} and BLEU~\citep{papineni2002bleu} as evaluation metrics. 

Additionally,  FEQA~\citep{durmus2020feqa} is used as a metric to evaluate the faithfulness of the answers. Given a summary of a source passage, FEQA masks the important tokens, which are extracted automatically by a constituency parser and a NER (Named Entity Recognition) model. Then, a learned model is used to generate questions corresponding to the masked tokens, and the masked tokens are considered as their ground truth answers. After that, the QA model replies to the generated question based on the source passage. Finally, FEQA measures the F1 score of the generated answers against the ground truth answers. It can be formulated as:
\begin{gather}
    a^g = QA \, Model(q^g, s)\\
    FEQA = F1-Score(a, a^g)
\end{gather}
where $q^g$ represents the generated question, $a^g$ is the generated answer corresponding to $q^g$, $s$ denotes the source passage and $a$ denotes the ground truth answer. In this paper, the dataset we use has already had the question-answer pairs (formulated as $q^d,a^d$). Hence, we do not generate the questions $q^g$, but instead use the existing questions $q^d$ in the dataset and treat the corresponding answers $a^d$ as the ground truth answers, which can be written as:
\begin{gather}
    a^g = QA \, Model(q^d, s)\\
    FEQA = F1-Score(a^d, a^g)
\end{gather}

\subsubsection{Hyperparameter Setup}
We use AdamW algorithm with learning rate $5\times10^{-4}$ to train the network. All experiments are conducted at $2\times$NVIDIA RTX A6000. The training epoch is set to $2$, the batch size per device is $1$, and the gradient accumulation steps are set to $64$. The dropout rate of GAT is $0.1$. The number of layers $L$ is $2$, and the number of heads $H$ is $8$. The weight of residual $\lambda$ is set to $0.2$. We adopt low-rank adaptation of large language models (LoRA)~\citep{hu2022lora} to fine-tune the parameters of foundation models. For each model, we run three times of the experiments with different random seeds and calculate the average results.

\subsection{Adaptability on Different Foundation LLMs}
In this part, we plug MALM into a range of different LLMs to test how much performance improvement it can achieve. This experiment is conducted on the HaluEval dataset and TruthfulQA dataset.
\subsubsection{Baselines}
To verify the adaptability of MALM on different types of LLMs, we choose five typical LLMs as the baselines:\\
\indent\textbf{GPT-2}~\citep{radford2019language}, which is trained on a large English corpus with self-supervised learning. That is, it learns to predict the next token in the sentence.\\
\indent\textbf{BLOOM}~\citep{workshop2022bloom}, which is a publicly released multilingual LLM, trained on 46 natural languages and 13 programming languages.\\
\indent\textbf{Qwen-2}~\citep{yang2024qwen2technicalreport} \textbf{\& Qwen-2.5}~\citep{qwen2025qwen25technicalreport}, which are Alibaba Cloud's general-purpose AI models. We use Qwen-2-7B and Qwen-2.5-7B as the baselines.\\
\indent\textbf{Falcon}~\citep{almazrouei2023falcon}, which is a causal decoder-only model trained on large-scale web data. Its architecture is based on PaLM. We use Falcon-7B as a baseline.\\
\indent\textbf{Vicuna}~\citep{zheng2023judging} is fine-tuned from LLaMA-2 on conversations from ShareGPT. We use Vicuna-7B-v1.5 as the baseline.\\
\indent\textbf{LLaMA-2}~\citep{touvron2023llama2}, which is released by Meta AI with diverse scales from 7B to 70B. We use LLaMA-2-7B as a baseline.\\

All the above baseline models are also fine-tuned by using LoRA and the learning rate is $5\times10^{-3}$ for GPT-2 and $5\times10^{-4}$ for the others. For each baseline model, we plugin MALM as adapter and get the result, which is then compared with the model's original result.

\subsubsection{Experimental Results and Analysis}
\begin{figure*}[t]
\centering
\includegraphics[width=1.0\textwidth]{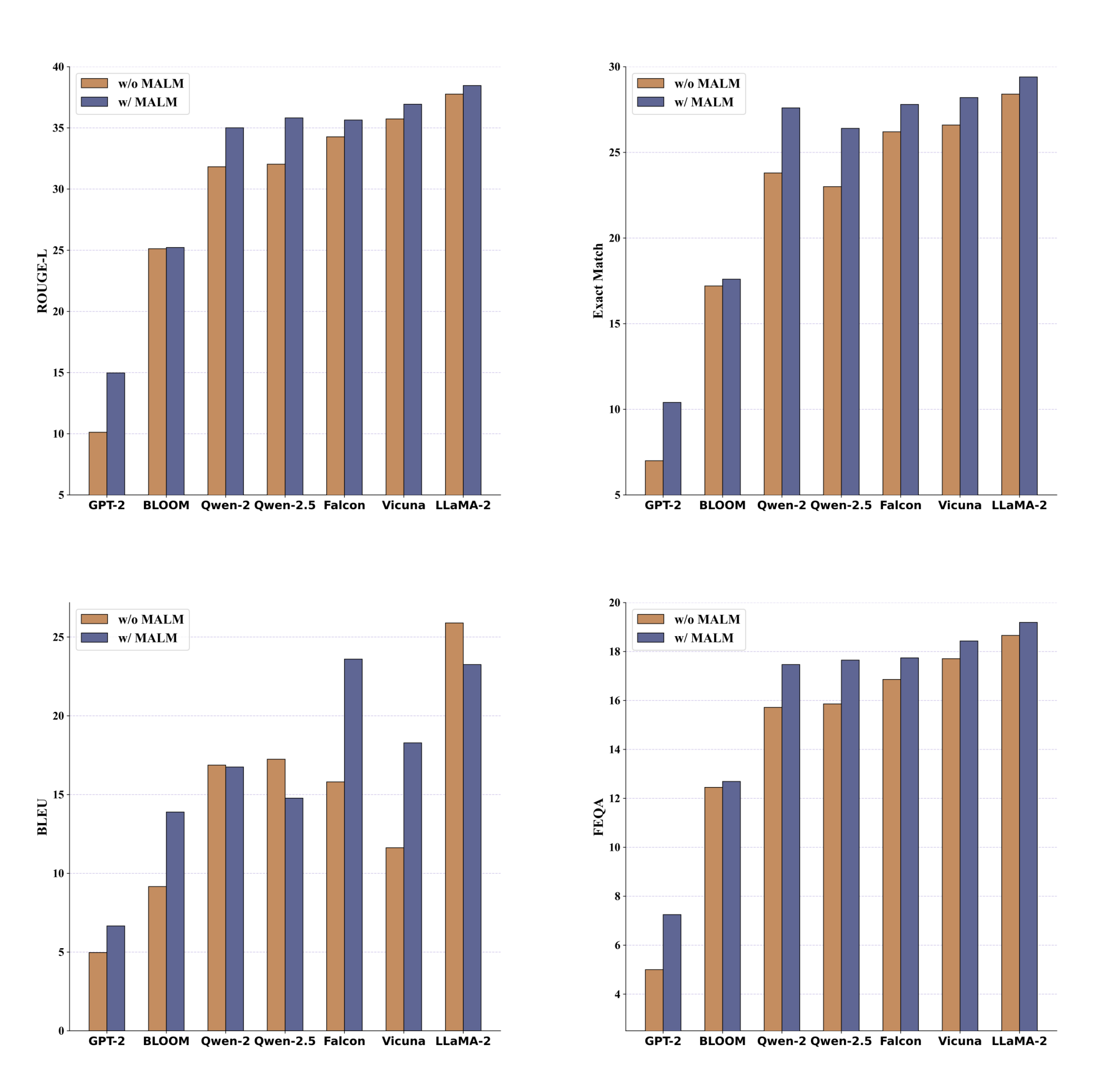}
\caption{The performance of foundation models w/o and w/ MALM.} \label{bar}
\end{figure*}

\textbf{Results on HaluEval.} The experimental results are shown in Table~\ref{tab:result}. For each baseline model, we test its fine-tuned performance without MALM, as well as the performance after incorporating MALM as adapter. It can be observed that MALM helps improve the performance of each baseline model. The results of foundation models w/o and w/ MALM are depicted in Fig.~\ref{bar}, where the performance of models is generally ranked in increasing order across each metric. Among them, GPT-2 has the lowest scores on all metrics. This is expected because it has the fewest parameters (124M) among all the baselines, while others have 7B parameters. At the same parameter level, BLOOM performs poorly. It may be due to BLOOM being a multilingual LLM, leading to insufficient pretraining data specifically in English, which is 485B Bytes. As a comparison, Falcon uses 1.5T tokens, LLaMA-2 uses 2T tokens, Qwen-2 uses 7T tokens, and Qwen-2.5 uses 18T tokens for pretraining. Vicuna is fine-tuned from LLaMA-2 by using conversations between ChatGPT and users. However, ChatGPT itself may generate responses with hallucination, and users might provide incorrect information as well. This could potentially affect the performance of Vicuna in terms of tackling the hallucination problem. LLaMA-2 outperforms all the other baselines. We think this result owes to the large pretraining corpus and the RLHF mechanism.

\begin{table*}[htb]
\caption{Experimental results in HaluEval. Each baseline model is fine-tuned with and without MALM (w/ represents 'with' and w/o represents 'without'). The results demonstrate that MALM can effectively improve the performance of all baselines, which means it has the potential to be adapted to mitigate hallucination in different LLMs.}
\label{tab:result}
\begin{center}
\scalebox{1.0}{
\begin{tabular}{c|c|cccccc}
\midrule[1pt]
\textbf{Models}          & \textbf{MALM} & \textbf{ROUGE-1} & \textbf{ROUGE-2} & \textbf{ROUGE-L} & \textbf{Exact Match} & \textbf{BLEU}  & \textbf{FEQA}  \\ \hline
\multirow{2}{*}{GPT-2}   & w/o           & 10.23            & 2.96             & 10.13            & 7.00                & 4.97           & 5.00           \\
                         & w/             & 14.90            & 4.38             & 14.97            & 10.40                & 6.66           & 7.25           \\ \hline
\multirow{2}{*}{BLOOM}   & w/o           & 25.23            & 10.05            & 25.13            & 17.20                & 9.16           & 12.45          \\
                         & w/             & 25.33            & 10.56            & 25.23            & 17.60                & 13.89          & 12.69          \\ \hline
\multirow{2}{*}{Qwen-2}   & w/o           & 32.14            & 15.05            & 31.83            & 23.80                & 16.87          & 15.72          \\
                         & w/             & 35.50            & 16.20            & 35.01            & 27.60                & 16.75          & 17.47          \\ \hline
\multirow{2}{*}{Qwen-2.5} & w/o           & 31.93            & 14.70            & 32.04            & 23.00                & 17.24          & 15.86          \\
                         & w/             & 35.88            & 16.95            & 35.82            & 26.40                & 14.77          & 17.65          \\ \hline
\multirow{2}{*}{Falcon}  & w/o           & 34.51            & 14.66            & 34.27            & 26.20                & 15.80          & 16.86          \\
                         & w/             & 35.63            & 16.13            & 35.65            & 27.80                & 23.60          & 17.74          \\ \hline
\multirow{2}{*}{Vicuna}  & w/o           & 35.77            & 15.78            & 35.74            & 26.60                & 11.62          & 17.71          \\
                         & w/             & 37.18            & 17.08            & 36.94            & 28.20                & 18.28          & 18.43          \\ \hline
\multirow{2}{*}{LLaMA-2} & w/o           & 37.91            & 17.86            & 37.76            & 28.40                & 25.89 & 18.66          \\
                         & w/             & 38.69   & 18.77   & 38.46   & 29.40       & 23.25          & 19.19 \\ \midrule[1pt]
\end{tabular}
}

\end{center}
\end{table*}

The proposed MALM has improved the performance of all foundation LLMs, and LLaMA-2 w/ MALM achieves the best scores of ROUGE-1, ROUGE-2, ROUGE-L, Exact Match and FEQA, which surpass LLaMA-2 by $2.05\%$, $5.10\%$, $1.85\%$, $3.52\%$ and $2.84\%$ respectively. All the improvements over the best baseline are statistically significant (two-tailed t-test, p-value $= 0.004 < 0.01$). We attribute the results to the incorporation of three types of connections, which we encode in the graph architecture to mitigate hallucination from a comprehensive perspective.

We can also notice that the BLEU score of LLaMA-2 overcomes LLaMA-2 w/ MALM. To uncover the reason, we investigate the brevity penalty of LLaMA-2, which is $0.86$. For comparison,  it is $0.91$ for LLaMA-2 w/ MALM. This indicates that the average length of generated text of LLaMA-2 is relatively short, which is conducive to obtain higher n-gram matching scores in BLEU. Furthermore, BLEU focuses on measuring exact matching degree between texts rather than assessing completeness and coverage level of information. This probably makes it less sensitive for evaluating generative tasks.

\textbf{Results on TruthfulQA.} Table~\ref{tab:result_truthfulqa} shows the results conducted on TruthfulQA. We can observe that MALM also effectively helps all models to improve their performance in the TruthfulQA dataset. Compared to the experimental results in the HaluEval dataset, the first difference we can notice is that the Qwen series models outperform other LLMs in all metrics, and Qwen-2 w/ MALM performs the best in terms of ROUGE-1 ($51.94\%$), ROUGE-2 ($37.81\%$), ROUGE-L ($49.35\%$), BLEU ($35.40\%$), and FEQA ($24.22\%$) metrics. We think it might be attributed to the mixture-of-experts (MoE) models built in the Qwen series models. MoE feed-forward neural networks (FFNs) replace the original FFNs in the Transformers block. Each MoE FFN is served as an individual expert. Facilitated by the expert combinations, Qwen reduces its possibility of making mistakes. And compared to the HaluEval dataset, questions in the TruthfulQA dataset are designed to make models more prone to falsehood, so Qwen's performance on the TruthfulQA dataset is more outstanding. However, Qwen-2.5 performs worse than Qwen-2. We consider this phenomenon might be due to overfitting. Qwen-2.5 has learned more complex factual knowledge through larger scale data during pre-training, while the training samples in the TruthfulQA dataset are relatively limited. When fine-tuned by LoRA in the same configuration as other baselines, Qwen-2.5 may overfit these limited samples, which actually damages its original knowledge expression. As for LLaMA-2, the improvement by MALM is more obvious: $85.94\%$ in ROUGE-1, $132.01\%$ in ROUGE-2, $87.61\%$ in ROUGE-L, $86.90\%$ in Exact Match, $29.28\%$ in BLEU, and $80.63\%$ in FEQA, which proves the effectiveness of MALM.

\begin{table*}[htb]
\caption{Experimental results in TruthfulQA. Each baseline model is fine-tuned with and without MALM (w/ represents 'with' and w/o represents 'without'). The results demonstrate that MALM can effectively improve the performance of all baselines, which means it has the potential to be adapted to mitigate hallucination in different LLMs.}
\label{tab:result_truthfulqa}
\begin{center}
\scalebox{1.0}{
\begin{tabular}{c|c|cccccc}
\midrule[1pt]
\textbf{Models}          & \textbf{MALM} & \textbf{ROUGE-1} & \textbf{ROUGE-2} & \textbf{ROUGE-L} & \textbf{Exact Match} & \textbf{BLEU}  & \textbf{FEQA}  \\ \hline
\multirow{2}{*}{GPT-2}   & w/o           & 24.87            & 13.01             & 22.86            & 0.01                & 4.74           & 9.42           \\
                         & w/             & 37.42            & 24.14             & 34.46            & 0.14                & 11.33           & 13.81           \\ \hline
\multirow{2}{*}{BLOOM}   & w/o           & 19.91            & 10.45            & 18.81            & 2.52                & 4.77           & 8.77          \\
                         & w/             & 39.32            & 27.37             & 37.52            & 3.28                & 13.05           & 15.55           \\ \hline
\multirow{2}{*}{Qwen-2}   & w/o           & 50.49            & 36.48            & 47.72            & 5.66                & 31.87          & 23.27          \\
                         & w/             & 51.94            & 37.81            & 49.35            & 7.55                & 35.40          & 24.22          \\ \hline
\multirow{2}{*}{Qwen-2.5} & w/o           & 45.38            & 30.57            & 43.04            & 6.29                & 27.90          & 21.28          \\
                         & w/             & 49.70            & 34.63            & 47.02            & 8.18                & 32.24          & 22.90          \\ \hline
\multirow{2}{*}{Falcon}  & w/o           & 24.42            & 14.38            & 22.59            & 2.52                & 13.15          & 10.40          \\
                         & w/             & 40.46            & 28.26            & 38.51            & 4.12                & 14.29          & 17.01          \\ \hline
\multirow{2}{*}{Vicuna}  & w/o           & 36.43            & 22.88            & 33.78            & 3.77                & 14.50          & 16.14          \\
                         & w/             & 38.38            & 26.40            & 36.49            & 4.90                & 12.85          & 17.07          \\ \hline
\multirow{2}{*}{LLaMA-2} & w/o           & 20.20            & 10.84            & 18.56            & 2.52                & 9.29          & 9.14          \\
                         & w/             & 37.56   & 25.15   & 34.82   & 4.71       & 12.01          & 16.51 \\ \midrule[1pt]
\end{tabular}
}

\end{center}
\end{table*}

\subsection{Generalizability to RAG}
Retrieval-augmented Generation (RAG)  combines retrieval-based and generative techniques to  tackle the hallucination problem caused by missing specific knowledge. Unlike traditional language models that rely solely on generating text based on patterns learned from large datasets, RAG incorporates a retrieval mechanism that accesses a vast knowledge base (or a document collection)  to enhance the LLM's responses. RAG has demonstrated significant advancements in various NLP tasks, including question answering, dialogue generation, and content summarization.

\subsubsection{Baselines}
We select four different models as the baselines, which are:\\
\indent\textbf{LLaMA-1}~\citep{touvron2023llama} is an open-source LLM, with its code and pretraining data  publicly available. The scale ranges from 7B to 65B, and we use LLaMA-7B as a baseline, due to computation constraints.\\
\indent\textbf{LLaMA-2}~\citep{touvron2023llama2} is an updated version of LLaMA-1 with 40\% increase in the size of pretraining corpus. We still use LLaMA-2-7B in this experiment.\\
\indent\textbf{SELF-RAG}~\citep{asai2024selfrag} is a state-of-the-art framework for RAG. When generating, it determines whether it is necessary to retrieve, evaluates the relevance of the retrieved passages, and also criticizes the generated output. This process enhances the generation factuality and quality.\\
\indent\textbf{RECOMP}~\citep{xu2023recomp}, another state-of-the-art RAG method, compresses the retrieved documents before inputting them into an LLM. We replace the Flan-UL2-20B in the original model to the LLaMA-2-7B for the sake of equity in comparison.

For each model introduced above, we combined three different retriever models respectively with them in the experiments:\\
\indent\textbf{BM25}~\citep{robertson2009probabilistic} is a classic  information retrieval model,  optimized from the TF-IDF algorithm.\\
\indent\textbf{Spider}~\citep{ram2022learning} is a dense retriever model, which is pretrained via self-supervised contrastive learning.\\
\indent\textbf{DPR}~\citep{karpukhin2020dense} uses dense vector representation to retrieve passages. The method optimizes the embedding of question and passage vectors by maximizing their inner products to avoid additional pretraining.

We also combine these three retriever models with LLaMA-2 w/ MALM to explore how effective MALM would be when serving as a RAG method.

\subsubsection{Experimental Results and Analysis}
We conduct this experiment on HaluEval, NQ and TriviaQA datasets.

\begin{table*}[htp]
\caption{Results of MALM generalized in RAG.}
\label{tab:rag}
\begin{center}
\scalebox{0.97}{
\begin{tabular}{ccccccccc}
\midrule[1pt]
Dataset & Model & Retriever & ROUGE-1 & ROUGE-2 & ROUGE-L & Exact Match & BLEU  & FEQA  \\ \hline
\multirow{15}{*}{HaluEval}
& \multirow{3}{*}{LLaMA-1}
& BM25      & 39.68   & 22.43   & 39.79   & 29.20       & 13.15 & 18.69 \\
&                                    
& Spider    & 38.66   & 21.45   & 38.33   & 28.60       & 12.81 & 17.76 \\
&                                    
& DPR       & 33.83   & 18.60   & 33.87   & 24.80       & 10.54 & 16.03 \\ \cline{2-9}
& \multirow{3}{*}{LLaMA-2}
& BM25      & 41.02   & 21.98   & 40.88   & 29.80       & 15.94 & 18.41 \\
&                                    
& Spider    & 37.82   & 19.97   & 37.80   & 27.00       & 13.02 & 17.40 \\
&                                    
& DPR       & 29.27   & 15.26   & 29.28   & 18.80       & 8.21  & 13.32 \\ \cline{2-9}
& \multirow{3}{*}{SELF-RAG}
& BM25      & 27.38   & 14.25   & 27.39   & 14.20       & 2.87  & 12.59 \\
&                                    
& Spider    & 25.53   & 14.76   & 25.46   & 13.00       & 2.65  & 11.73 \\
&                                    
& DPR       & 22.11   & 11.82   & 21.90   & 10.20       & 2.04  & 10.25 \\ \cline{2-9}
& \multirow{3}{*}{RECOMP}
& BM25      & 44.04   & 22.20   & 43.93   & 34.00       & 25.18 & 19.84 \\
&                                    
& Spider    & 38.59   & 20.24   & 38.55   & 29.20       & 25.63 & 17.15 \\
&                                    
& DPR       & 34.93   & 16.53   & 34.82   & 27.00       & 15.38 & 15.33 \\ \cline{2-9}
& \multirow{3}{*}{LLaMA-2 w/ MALM}
& BM25      & \textbf{51.78}   & \textbf{27.43}   & \textbf{51.54}   & \textbf{42.00}       & \textbf{37.66} & \textbf{25.38} \\
&                                    
& Spider    & 47.01   & 24.04   & 46.84   & 38.60       & 33.33 & 23.19 \\
&                                    
& DPR       & 40.63   & 20.77   & 40.40   & 32.00       & 23.49 & 20.10 \\ \hline
\multirow{15}{*}{NQ}
& \multirow{3}{*}{LLaMA-1}
& BM25      & 25.72   & 11.98   & 25.58   & 13.66       & 4.03  & 10.92 \\
&                                    
& Spider    & 26.83   & 12.07   & 26.61   & 14.76       & 4.14  & 11.57 \\
&                                    
& DPR       & 32.10   & 16.09   & 31.97   & 18.03       & 5.90  & 14.01 \\ \cline{2-9}
& \multirow{3}{*}{LLaMA-2}
& BM25      & 21.13   & 9.79    & 20.96   & 10.53       & 2.45  & 9.20  \\
&                                    
& Spider    & 22.76   & 10.94   & 22.62   & 11.72       & 2.83  & 9.78  \\
&                                    
& DPR       & 27.57   & 14.31   & 27.44   & 14.60       & 3.89  & 12.17 \\ \cline{2-9}
& \multirow{3}{*}{SELF-RAG}          
& BM25      & 31.41   & 17.01   & 31.32   & 20.89       & 12.60 & 15.06 \\
&                                    
& Spider    & 32.05   & 17.37   & 31.96   & 21.61       & 13.39 & 15.49 \\
&                                    
& DPR       & 35.80   & 19.10   & 35.65   & \textbf{21.05}       & \textbf{16.38} & 16.75 \\ \cline{2-9}
& \multirow{3}{*}{RECOMP}            
& BM25      & 30.31   & 14.24   & 30.02   & 13.21       & 2.81  & 11.68 \\
&                                    
& Spider    & 30.05   & 14.72   & 29.64   & 13.16       & 2.99  & 11.50 \\
&                                    
& DPR       & 34.70   & 17.72   & 34.34   & 15.98       & 3.90  & 13.48 \\ \cline{2-9}
& \multirow{3}{*}{LLaMA-2 w/ MALM}
& BM25      & 31.36   & 16.12   & 31.24   & 18.28       & 11.86 & 14.59 \\
&                                    
& Spider    & 32.29   & 16.84   & 32.15   & 18.56       & 12.86 & 15.09 \\
&                                    
& DPR       & \textbf{37.49}   & \textbf{21.40}   & \textbf{37.31}   & 20.66       & 11.88 & \textbf{17.54} \\ \hline
\multirow{15}{*}{TriviaQA}
& \multirow{3}{*}{LLaMA-1}
& BM25      & 33.12   & 13.01   & 32.57   & 11.99       & 7.51  & 11.19 \\
&                                    
& Spider    & 31.68   & 12.30   & 31.14   & 11.17       & 6.98  & 10.71 \\
&                                    
& DPR       & 32.98   & 12.89   & 32.47   & 11.75       & 7.27  & 11.27 \\ \cline{2-9}
& \multirow{3}{*}{LLaMA-2}
& BM25      & 32.38   & 12.90   & 31.88   & 11.38       & 6.44  & 11.00 \\
&                                    
& Spider    & 30.83   & 12.16   & 30.31   & 10.78       & 6.02  & 10.52 \\
&                                    
& DPR       & 32.24   & 12.50   & 31.72   & 11.31       & 6.48  & 10.64 \\ \cline{2-9}
& \multirow{3}{*}{SELF-RAG}          
& BM25      & 20.13   & 8.84    & 19.60   & 4.52        & 1.12  & 7.07  \\
&                                    
& Spider    & 19.01   & 8.06    & 18.47   & 4.23        & 1.05  & 6.63  \\
&                                    
& DPR       & 19.34   & 8.27    & 18.80   & 4.39        & 1.07  & 6.76  \\ \cline{2-9}
& \multirow{3}{*}{RECOMP}            
& BM25      & 33.20   & 12.62   & 32.70   & 11.87       & 8.19  & 11.06 \\
&                                    
& Spider    & 31.92   & 11.94   & 31.40   & 11.39       & 8.21  & 10.65 \\
&                                    
& DPR       & 32.98   & 12.44   & 32.51   & 11.92       & 8.25  & 11.02 \\ \cline{2-9}
& \multirow{3}{*}{LLaMA-2 w/ MALM} 
& BM25      & \textbf{33.80}   & \textbf{13.11}   & \textbf{33.26}   & \textbf{12.22}       & \textbf{9.30}  & \textbf{11.28} \\
&                                    
& Spider    & 32.27   & 12.29   & 31.76   & 11.77       & 8.34  & 10.87 \\
&                                    
& DPR       & 33.29   & 12.80   & 32.77   & 12.00       & 7.84  & 11.07 \\ \midrule[1pt]
\end{tabular}
    }
\end{center}
\end{table*}

\textbf{Results on HaluEval.} As shown in Table~\ref{tab:rag}, we can observe that, without using MALM, the DPR retriever performs the worst, while BM25  performs the best for all of the models. The reason may be that DPR is a dense retriever which needs to be trained on the corpus to maximize its potential. However, it is not trained on this new dataset for hallucination mitigation in our experiment. Among the baseline models, SELF-RAG achieves the least good generation performance for hallucination mitigation in terms of all metrics. We can deduce that although SELF-RAG has previously achieved good results in RAG tasks, there is still a long way for improvement in hallucination mitigation. LLaMA-1 and LLaMA-2 performs better than SELF-RAG. After observing the results of LLaMA-1 and LLaMA-2 with different retrievers, it can be found that their performance is about equal. LLaMA-2 with BM25 performs better than LLaMA-1 with BM25 in terms of ROUGE-1, ROUGE-L, Exact Match and BLEU, while LLaMA-1 with BM25 outperforms LLaMA-2 with BM25 in terms of ROUGE-2 and FEQA. Compared to LLaMA-1, the performance of LLaMA-2 does not show an obvious improvement, which demonstrates that LLaMA-2 still suffers from the hallucination problem even though the pretraining corpus is expanded. RECOMP achieves the best performance among the baseline models. This result may attribute to the information density of the documents that RECOMP retrieves and compresses. When using BM25 as retriever, LLaMA-2 w/ MALM produces a new state-of-the-art performance, with improvements by $17.57\%$, $23.56\%$, $17.32\%$, $23.53\%$, $49.56\%$ and $27.92\%$ over RECOMP in terms of ROUGE-1, ROUGE-2, ROUGE-L, Exact Match, BLEU and FEQA respectively. All the improvements over the best baseline are statistically significant (two-tailed t-test, p-value $= 0.002 < 0.01$). This indicates that MALM can yield better responses when used as RAG approach. By leveraging knowledge retrieved by RAG, MALM better utilizes the adapter's function, guiding and assisting in generating responses with lower hallucination within the LLM. This is not only attributable to the assistance provided by the external knowledge retrieved for the generation, but also to MALM's comprehensive consideration of input information, context information, and knowledge information.

\textbf{Results on NQ.} NQ contains more question-answer pairs from real human search, and the extensiveness of this dataset can further test the model's generalizability. Comparing to the results of HaluEval, we can observe that the performance of most models has decreased visibly. This proves it is challenging to generalize models to a larger dataset. LLaMA-1 and LLaMA-2 perform poorly in terms of ROUGE, since they are single language models without extra support designed for hallucination mitigation. However, we can observe an abnormal phenomenon that the results of LLaMA-2 are worse than LLaMA-1 with the same retriever in terms of all the metrics. The reason may be that the knowledge information retrieved by the retriever cannot be guaranteed to be completely accurate. When using such knowledge information as prompt input to the model, LLaMA-2 is more susceptible to hallucination, because it is more sensitive to the context than LLaMA-1 and there is no other mechanism to mitigate this influence. RECOMP performs better than LLaMA-1 and LLaMA-2 in terms of ROUGE. SELF-RAG achieves the best performance among all baselines in terms of all metrics.  LLaMA-2 w/ MALM surpasses SELF-RAG in most metrics, but it under-performs SELF-RAG in Exact Match and BLEU metrics. We argue that these two metrics only consider the degree of positional matching of word sequences, neglecting the matching at the semantic level.  LLaMA-2 w/ MALM may generate sentences that are semantically consistent with but differ from ground truth in terms of word order to ensure the naturalness and coherence of the generation. This indeed can be proved by fact that LLaMA-2 w/ MALM has achieved the state-of-the-art results of $37.49\%$, $21.40\%$, $37.31\%$, $17.54\%$ in terms of ROUGE-1, ROUGE-2, ROUGE-L and FEQA. All the improvements over the best baseline are statistically significant (two-tailed t-test, p-value $= 0.003 < 0.01$).

\textbf{Results on TriviaQA.} The results on TriviaQA are in general lower than those on HaluEval, yet showing the same trend. The performance of SELF-RAG is the worst. TriviaQA has complex questions, and requires the models to have the ability in cross sentence reasoning, which may be what SELF-RAG lacks. RECOMP performs slightly better than LLaMA-1 and LLaMA-2. The compressing process in RECOMP may help overcome the difficulty in finding answers across sentences. LLaMA-2 w/ MALM outperforms all the baselines, and achieves the state-of-the-art performance with a ROUGE-1 score of $33.80\%$, ROUGE-2 of $13.11\%$, ROUGE-L of $33.26\%$, Exact Match of $12.22\%$, BLEU of $9.30\%$, and FEQA of $11.28\%$. All the improvements over the best baseline are statistically significant (two-tailed t-test, p-value $= 0.006 < 0.01$). These results demonstrate that the ability of MALM to mitigate hallucination can also play a role in reading comprehension tasks, and prove the generalizability of the proposed MALM on different tasks.

\begin{table*}[htp]
\caption{Top-k retrieval accuracy on the test sets of three datasets.}
\label{tab:retrieval}
\begin{center}
\scalebox{1.0}{
\begin{tabular}{ccccccc}
\midrule[1pt]
Dataset                   & Retriever & Top-1          & Top-5          & Top-20         & Top-50         & Top-100        \\ \hline
\multirow{3}{*}{HaluEval} & BM25      & \textbf{34.40} & \textbf{51.80} & \textbf{65.00} & \textbf{71.60} & \textbf{73.80} \\
                          & spider    & 26.20          & 40.20          & 55.40          & 64.40          & 69.80          \\
                          & DPR       & 23.00          & 34.80          & 51.80          & 59.40          & 65.00          \\ \hline
\multirow{3}{*}{NQ}       & BM25      & 22.11          & 43.77          & 62.99          & 72.74          & 78.23          \\
                          & spider    & 24.76          & 49.56          & 68.20          & 76.90          & 81.19          \\
                          & DPR       & \textbf{44.40} & \textbf{67.04} & \textbf{79.47} & \textbf{83.80} & \textbf{86.09} \\ \hline
\multirow{3}{*}{TriviaQA} & BM25      & 46.30          & 66.29          & 76.41          & 80.58          & 83.14          \\
                          & spider    & 41.77          & 63.64          & 75.75          & 80.40          & 83.47          \\
                          & DPR       & \textbf{52.78} & \textbf{69.82} & \textbf{78.87} & \textbf{82.59} & \textbf{84.79} \\ \midrule[1pt]
\end{tabular}
    }
\end{center}
\end{table*}

\textbf{Effect of Retrievers.} We also evaluate the top-k retrieval accuracy of the three retrievers, i.e. the percentage of questions for which the answer span is found in the top-k passages, as shown in Table.~\ref{tab:retrieval}. On the HaluEval dataset, BM25 performs the best, while DPR performs the best on the NQ dataset. These results correspond exactly to the results in Table.~\ref{tab:rag}. That is, LLaMA-2 w/ MALM achieves the best performance on HaluEval when using BM25 and achieves the best performance on NQ when using DPR as the retriever. However, DPR is inferior to BM25 for LLaMA-2 w/ MALM on TriviaQA dataset, though it reaches the highest accuracy in Table.~\ref{tab:retrieval}. This phenomenon may be due to the following reasons: the difficulty in utilizing the retrieved passages is relatively high, the model is not aware of the answers contained in the passages, and redundant information may mislead the model.

\textbf{Dataset Characteristics Analysis} To systematically investigate how dataset attributes influence MALM's performance (for consistency, we only compare the results of MALM with BM25), we conduct a multi-dimensional comparison across HaluEval, NQ, and TriviaQA, which is summarized in Table.~\ref{tab:dataset_analysis}. Key findings include: (1) Question Type: We count interrogative words across all questions in the datasets and present the top three most frequent types in Table.~\ref{tab:dataset_analysis}. Across all evaluation metrics, MALM achieves superior performance on the HaluEval dataset. This can be attributed to the prevalence of What/Which-type questions (44\% \& 32\%) that focus on entity identification, a task particularly suited to MALM's graph architecture, which effectively aligns answers with the information of entities in knowledge. In contrast, despite TriviaQA's similar dominance of Which/What questions (43\% \& 36\%), MALM exhibits reduced effectiveness (e.g., 12.22\% Exact Match vs. HaluEval's 42.00\%). We hypothesize this discrepancy stems from stronger multi-hop reasoning demand in TriviaQA. The options in Which-type questions are often puzzling, requiring reasoning capability that MALM's current architecture lacks explicit mechanisms to address; (2) Knowledge Domain: Given that the knowledge domains of all three datasets (HaluEval, NQ, TriviaQA) in this experiment are exclusively sourced from Wikipedia, the variations of MALM's performance are not related to the knowledge domain characteristics of datasets; (3) Length Sensitivity: MALM's performance is positively correlated with the average question length - longer questions lead to better performance, indicating that longer questions provide richer context information, and thus MALM's performance improves. On the other hand, it is negatively correlated with the average knowledge length, suggesting that longer knowledge tokens reduce MALM's information extraction efficiency; (4) Structure: Although the structures of different datasets vary, in this experiment, we standardize them into the same format as HaluEval (QA pairs \& Knowledge) before use, ensuring that dataset structures do not affect MALM's performance.

\begin{table*}[htp]
\caption{Critical distinctions between the benchmark datasets.}
\label{tab:dataset_analysis}
\begin{center}
\scalebox{1.0}{
\begin{tabular}{cccc}
\midrule[1pt]
\textbf{Dataset}                         & HaluEval              & NQ                                    & TriviaQA                \\ \hline
\multirow{3}{*}{\textbf{Question Type}} & What (44\%)           & Who (37\%)                            & Which (43\%)            \\
                                         & Which (32\%)          & When (18\%)                           & What (36\%)             \\
                                         & Who (11\%)            & What (16\%)                           & Who (14\%)              \\ \hline
\textbf{Knowledge Domain}                & Wikipedia             & Wikipedia                             & Wikipedia \\ \hline
\textbf{Avg. Question Length}            & 27 tokens             & 14 tokens                             & 22 tokens               \\ 
\textbf{Avg. Knowledge Length}           & 99 tokens             & 166 tokens                            & 168 tokens              \\ \hline
\textbf{Structure}              & QA pairs \& Knowledge & Questions \& Documents \& Annotations & QA pairs \& Documents   \\ \midrule[1pt]
\end{tabular}
    }
\end{center}
\end{table*}

\subsection{Ablation Study}
In the following experiments, the foundation model of MALM is set as LLaMA-2. For the sake of brevity, we hereafter refer to it as MALM.

\begin{table}[h]
\caption{Ablation study results.}
\label{tab:ablation}
\begin{center}
\scalebox{1.0}{
\begin{tabular}{cccc}
\midrule[1pt] 
\textbf{Models} & \textbf{ROUGE-1} & \textbf{ROUGE-L} & \textbf{Exact Match} \\ \hline
w/o Context   & 36.84   & 36.63   & 28.00       \\
w/ Full Context  & 37.79   & 37.48   & 27.60       \\
w/o Input     & 36.71   & 36.50   & 26.40       \\
w/o Knowledge & 38.23   & 36.20   & 28.80       \\ \hline
MALM     & 38.69   & 38.46   & 29.40       \\ \midrule[1pt] 
\end{tabular}
    }
\end{center}
\end{table}

To verify the contributions of different components in MALM, we conduct ablation study on the HaluEval dataset with four variants:

(1) \textit{w/o Context}: the masked full connections among partial output nodes are eliminated to test the importance of context information;

(2) \textit{w/ Full Context}: the masked full context connections are replaced with full connections to verify the effectiveness of masked full connections;

(3) \textit{w/o Input}: the connections from input nodes towards partial output nodes are eliminated to test the importance of the input information;

(4) \textit{w/o Knowledge}: the connections from knowledge nodes towards partial output nodes are removed to observe how important the knowledge information is.

The main results are shown in Table~\ref{tab:ablation}. We can find that \textit{w/o Input} achieves the worst performance among all ablated models in terms of ROUGE-1 and Exact Match, which decreases by $5.12\%$ and $10.20\%$ respectively in comparison with MALM. It means that removing the input connections has the greatest influence on the performance of MALM to generate correct responses. That is, the input information attributes the most to the performance of hallucination mitigation. \textit{w/o Knowledge} is the worst ablated model in terms of ROUGE-L, which decreases by $5.88\%$ compared with MALM. It means that knowledge connections are also crucial to the overall performance, and knowledge information also plays an important role to mitigate hallucination. These results indicate that the main factors influencing MALM's performance are the structure of input connections and the knowledge information.

When comparing the results of \textit{w/o Context} and \textit{w/ Full Context}, we can observe that \textit{w/o Context} performs worse than \textit{w/ Full Context} in terms of ROUGE-1 and ROUGE-L. The result can demonstrate that the use of masked full connections outperforms the use of full connections in context connection, which proves the usefulness of the context connection structure of MALM.

\subsection{Effects of Model Layers}
MALM consists of multiple layers of graph networks, so that different information can be aggregated in the updating process through layers. To find out how the number of layers affects the final result, we vary the value of $L$ ranging from 0 to 4 in MALM and MALM with BM25, and test them on the HaluEval dataset. Note that $L=0$ is equivalent to the model without MALM. The results are shown in Table.~\ref{tab:layer}.

\begin{table}[htb]
\caption{Results of varying numbers of model layers.}
\label{tab:layer}
\begin{center}
\scalebox{1.0}{
\begin{tabular}{cccccccc}
\midrule[1pt] 
Model & Layer & ROUGE-1 & ROUGE-2 & ROUGE-L & Exact Match & BLEU  & FEQA  \\ \hline
\multirow{5}{*}{MALM}
& L=0   & 37.91   & 17.86   & 37.76   & 28.40       & 25.89 & 18.66 \\
& L=1   & 37.88   & 17.87   & 37.50   & 27.80       & 21.10 & 18.72 \\
& L=2   & 38.69   & 18.77   & 38.46   & 29.40       & 23.25 & 19.19 \\
& L=3   & 36.55   & 16.62   & 36.47   & 27.20       & 21.37 & 18.17 \\
& L=4   & 36.37   & 18.12   & 36.33   & 27.20       & 23.20 & 18.10 \\ \hline
\multirow{5}{*}{MALM with BM25}
& L=0   & 41.02   & 21.98   & 40.88   & 29.80       & 15.94 & 18.41 \\
& L=1   & 53.12   & 27.52   & 52.97   & 42.60       & 37.00 & 26.03 \\
& L=2   & 51.78   & 27.43   & 51.54   & 42.00       & 37.66 & 25.38 \\
& L=3   & 51.87   & 27.49   & 51.71   & 42.60       & 37.50 & 25.56 \\
& L=4   & 52.63   & 27.04   & 52.53   & 42.80       & 37.93 & 25.97 \\ \midrule[1pt] 
\end{tabular}
    }

\end{center}
\end{table}
\begin{figure*}[h]
\centering
\includegraphics[width=0.8\textwidth]{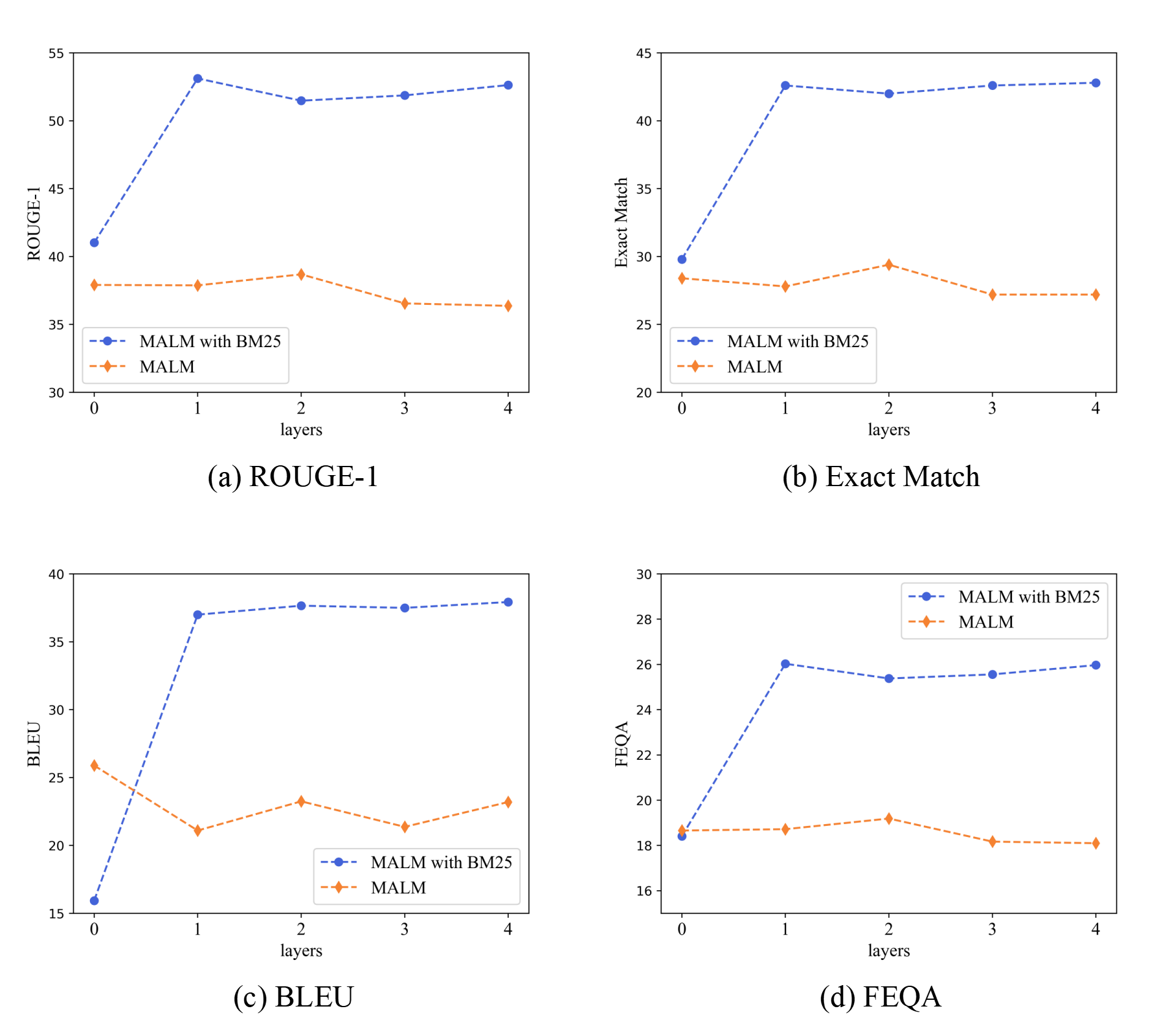}
\caption{Performance curves with respect to different layers.} \label{layer}
\end{figure*}

We can observe that the optimal number of the model layers is 2 for MALM. As shown in Fig.~\ref{layer}, when the number of layers grows from 0 to 2, the results generally show an increasing trend. This is because introducing the three types of graphs can help LLMs explore more relationships between input information, context information and knowledge information. The multi-layer structure increases the interactivity between different information in the model, improving the utilization of the information. However, the performance decreases as the number of layers continues to grow. The most likely reason is due to the over-smoothing problem. In multi-layer graph networks, the features of nodes at each layer are merged with their neighbors. Hence, the features of all nodes may converge to a similar state after multiple layers. It leads to a decrease in feature discrimination between different nodes, resulting in a reduction of performance. As for MALM with BM25, most shapes in Fig.~\ref{layer} approximate convex curves. The best performance is achieved at $L=1$ in terms of ROUGE and FEQA, and at $L=4$ in terms of Exact Match and BLEU. When the number of layers increases from 1 to 2, the performance decreases. However, the performance grows back when we continue to stack the layers. This phenomenon may be attributed to the fact that the average length of knowledge retrieved by BM25 (which is $631.92$) is relatively greater than that provided in HaluEval dataset (which is $342.97$). This leads to an increase in the number of nodes in the graph, elevating the complexity of the structure and making it more difficult for nodes to converge to a similar state. While maintaining feature discrimination, the interaction between different nodes has become more frequent.

\subsection{Evaluation with GPT-4}
In addition to the afore-reported empirical evaluation on relevant benchmarks, we also carry out  evaluations with external judges. We first employ GPT-4~\citep{openai2023gpt} as a judge. GPT-4 demonstrates that its performance in majority of professional and academic domains is close to human-level, and it also outperforms the existing SOTA LLMs on various benchmarks. Thus, GPT-4 is the most appropriate LLM as an automated evaluation judge. This evaluation is conducted on HaluEval dataset.

\begin{table}[h]
\caption{Instruction for automated evaluation.}
\label{tab:instruction}
\begin{center}
\scalebox{1.0}{
\begin{tabular}{p{0.4\textwidth}}
\midrule[1pt] 
\rowcolor[HTML]{FFCCC9}
As a hallucination identifier, I will provide you with a user query, relevant knowledge, corresponding ground truth, as well as option A and option B.\\
\rowcolor[HTML]{FFFC9E}
Option A and option B may contradict the user query or knowledge, and they may also contradict themselves. We refer to this phenomenon as hallucination.\\
\rowcolor[HTML]{96FFFB}
As a hallucination identifier, please select an option from A and B that is less prone to hallucination. If neither of them has hallucination, choose one that you think is more fluent and natural. Please only response with A or B. Let's begin.\\
Query: \textit{Style was the single that Taylor Swift released after which other song from her album 1989?}\\
Knowledge: \textit{"Blank Space" is a song by American singer-songwriter Taylor Swift from her fifth studio album "1989" (2014).}\\
Ground Truth: \textit{Blank Space}\\
Option A: \textit{Blank Space}\\
Option B: \textit{Shake It Off}\\ \midrule[1pt] 
\end{tabular}
    }
\end{center}
\end{table}

\begin{figure}[h]
\centering
\includegraphics[width=0.7\textwidth]{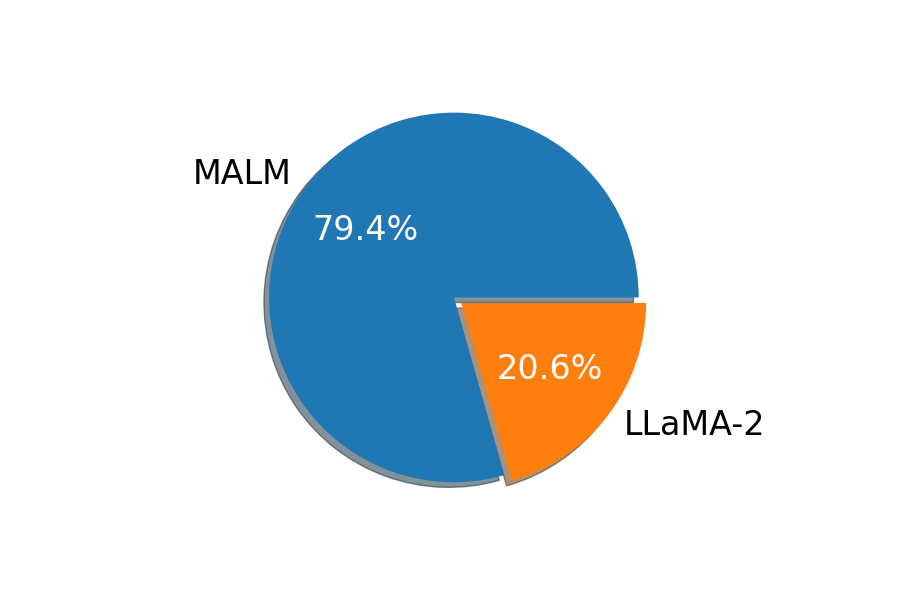}
\caption{The result of automated evaluation.} \label{pie}
\end{figure}

As shown in Table~\ref{tab:instruction}, we feed the instruction to GPT-4 to obtain the judgement. The red part is the preliminary which tells the role of GPT-4. The yellow part teaches GPT-4 the concept of hallucination.  The blue part states the detailed requirements. Option A is the response of MALM, and option B is the response of the best baseline, i.e., fine-tuned LLaMA-2. We adopt the whole test set to conduct automated evaluation, which contains 500 samples. We repeat the process three times to mitigate the impact of randomness, and take the average of the three evaluations as the final result, which is shown in Fig.~\ref{pie}.

We can see that the judge (i.e., GPT-4) favours MALM in all samples with a ratio of $79.4\%$. The result further proves that MALM can effectively mitigate hallucination in LLMs.

\subsection{Human Evaluation}
To further validate the effectiveness of MALM, we recruit three volunteers, who are postgraduate students in computer science and are familiar with LLMs, to evaluate the generated results. To this end, we sample 30 items from the test set, each of which contains a query, the relevant knowledge about the query, a ground truth answer, and two responses (one from MALM and the other from the fine-tuned LLaMA-2). The volunteers are unaware of which model the responses come from. Then, we teach them the three types of hallucination and how to distinguish them. The volunteers are well-educated, ensuring their comprehension of the hallucination problem. They are then required to determine which response is better given the query, the relevant knowledge and the ground truth.

\begin{table}[h]
\caption{Results of human evaluation.}
\label{tab:human}
\begin{center}
\scalebox{1.0}{
\begin{tabular}{ccc}
\midrule[1pt] 
\multicolumn{1}{l}{} & \textbf{MALM}     & \textbf{LLaMA-2}       \\ \hline
1                     & 21            & 9             \\
2                     & 19            & 11            \\
3                     & 19            & 11            \\ \hline
\textbf{Average}               & 19.7 (65.6\%) & 10.3 (34.4\%) \\ \midrule[1pt] 
\end{tabular}
    }
\end{center}
\end{table}

\begin{figure}[h]
\centering
\includegraphics[width=0.6\textwidth]{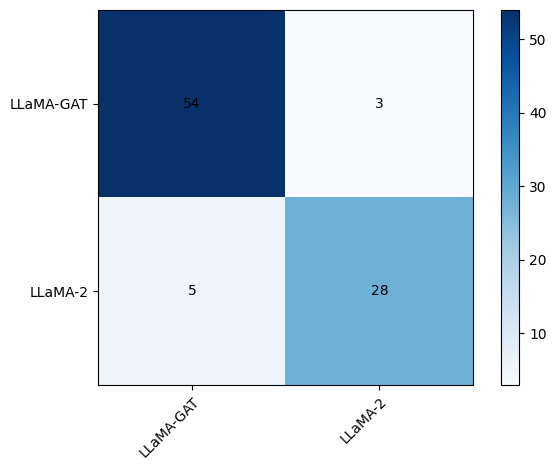}
\caption{The confusion matrix of human evaluation.} \label{matrix}
\end{figure}

Table~\ref{tab:human} shows the human evaluation results on the HaluEval dataset. The volunteers deem responses from MALM superior in $65.6\%$ of the samples. Finally, we adopt the max-voting strategy to determine the better response in each sample, and calculate inter-rater reliability by computing the Kappa scores. According to the confusion matrix in Fig.~\ref{matrix}, we can obtain the Kappa score of $0.81$, proving the high degree of agreement between the volunteers.

\subsection{Type-Specific Mitigation Analysis}
To evaluate MALM’s differential performance in mitigating distinct types of hallucination, we conduct an extra statistical analysis in this section to quantify MALM’s effects of different types of hallucination.

For each dataset, we first randomly selected 25 samples, resulting in a total of 100 evaluation samples from four datasets (HaluEval, TruthfulQA, NQ, and TriviaQA). Subsequently, we use both LLaMA-2 and MALM to generate answers for these 100 samples. Then we manually count the hallucinated answers generated by the two models and annotate their hallucination types. The final statistical results are presented in Table.~\ref{tab:statistic}, which compares the hallucination rates and type distributions of LLaMA-2 before and after being assisted by MALM.

\begin{table}[hbt]
\caption{Statistical results of type-specific mitigation analysis.}
\label{tab:statistic}
\begin{center}
\scalebox{1.0}{
\begin{tabular}{c|c|c|ccc|c}
\midrule[1pt]
\multirow{2}{*}{Model}          & \multirow{2}{*}{Dataset} & \multirow{2}{*}{Non-Hallucinated} & \multicolumn{3}{c|}{Hallucinated}                                     & Total \\ \cline{4-6}
                                &                          &                                   & \multicolumn{1}{c|}{Input} & \multicolumn{1}{c|}{Context} & Fact &       \\ \hline
\multirow{5}{*}{LLaMA-2} & HaluEval                 & 10                                  & \multicolumn{1}{c|}{2}      & \multicolumn{1}{c|}{0}        & 13          & 25    \\
                                & TruthfulQA               & 3                                  & \multicolumn{1}{c|}{7}      & \multicolumn{1}{c|}{2}        & 13          & 25    \\
                                & NQ                       & 12                               & \multicolumn{1}{c|}{2}      & \multicolumn{1}{c|}{1}        & 10          & 25    \\
                                & TriviaQA                 & 19                                  & \multicolumn{1}{c|}{0}      & \multicolumn{1}{c|}{0}        & 6          & 25    \\
                                & Total                    & 44                                  & \multicolumn{1}{c|}{11}      & \multicolumn{1}{c|}{3}        & 42          & 100   \\ \midrule[1pt]
\multirow{5}{*}{MALM}  & HaluEval                 & 12                                  & \multicolumn{1}{c|}{1}      & \multicolumn{1}{c|}{0}        & 12          & 25    \\
                                & TruthfulQA               & 15                                  & \multicolumn{1}{c|}{2}      & \multicolumn{1}{c|}{1}        & 7          & 25    \\
                                & NQ                       & 17                                  & \multicolumn{1}{c|}{0}      & \multicolumn{1}{c|}{0}        & 8          & 25    \\
                                & TriviaQA                 & 20                                  & \multicolumn{1}{c|}{0}      & \multicolumn{1}{c|}{0}        & 5          & 25    \\
                                & Total                    & 64                                  & \multicolumn{1}{c|}{3}      & \multicolumn{1}{c|}{1}        & 32          & 100   \\ \midrule[1pt]
\end{tabular}
    }
\end{center}
\end{table}

Overall, LLaMA-2 achieves a non-hallucinated answer rate of 44\%, while MALM improves the rate to 64\%, marking an enhancement of 20 percentage points, which demonstrates the high efficiency of MALM for hallucination mitigation. Among all the hallucinated answers generated by LLaMA-2, Input-Conflicting hallucination accounts for 19.6\%, Context-Conflicting hallucination accounts for 5.4\%, and Fact-Conflicting hallucination accounts for 75.0\%. This indicates that the most common type of hallucination is Fact-Conflicting hallucination in the experiments of our work, while the least common type is Context-Conflicting hallucination. But this does not mean that the Context-Conflicting hallucination is not important. Because the task of models in our experiments is question answering, where the length of the answers is often relatively short. Therefore the generating probability of Context-Conflicting hallucination is relatively low inherently in our experiments. With MALM integration, hallucination rates are reduced by 72.7\% for Input-Conflicting hallucination, 66.7\% for Context-Conflicting hallucination, and 23.8\% for Fact-Conflicting hallucination. This indicates that MALM can effectively mitigate all three types of hallucination and shows the strongest suppression effect on Input-Conflicting hallucination. We find that MALM can help models align with the input questions better, for example, when asked "What are the harms of sugar?", the original response "Sugar is a type of carbohydrate" is corrected by MALM to "Sugar may result in cavities". However, 32\% of Fact-Conflicting hallucination still persists with the assistance of MALM. We think this is due to the following two reasons: (1) The knowledge in the datasets or retrieved by the retrievers does not contain the correct answer. The main function of MALM is to help models better utilize knowledge information. If the knowledge information itself is not effective or abundant enough, MALM may not give full play to its function. After using MALM, models can learn to generate answers as much as possible based on the knowledge information, such as names that have not appeared in the questions but in the knowledge. However, if the correct answers are also not in the knowledge, models still cannot give the right answers; (2) There are noisy samples in the datasets. For instance, we find that a sample's question is "When did the Black Death end in England?", but the ground truth answer given by the dataset is "December 1349", which may refer to the beginning time of the Black Death in England rather than the end time. Fortunately, MALM has given the correct answer of "17th century". Noisy data leads to a decrease in the training effectiveness of models, therefore needs to be cleaned.

\subsection{Case Study}

\begin{table}[h]
\caption{Two samples for case study.}
\label{tab:case}
\begin{center}
\scalebox{1.0}{
\begin{tabular}{p{0.4\textwidth}}
\midrule[1pt] 
\textit{Case 1}:                                                                                                                                                   \\
\textbf{Query}: Which documentary film, Tender Fictions or Wheel of Time, is a 2003 documentary directed by German filmmaker Werner Herzog about Tibetan Buddhism? \\
\textbf{Knowledge}: Tender Fictions is a 1996 autobiographical documentary film directed by American experimental filmmaker Barbara Hammer.Wheel of Time is a 2003 documentary film by German director Werner Herzog about Tibetan Buddhism. \\
\textbf{Ground Truth}: Wheel of Time                                                                                                                                             \\
\textbf{Response of LLaMA-2}: Tender Fictions                                                                                                                                     \\
\textbf{Response of MALM}: Wheel of Time                                                                                                                                        \\\hline
\textit{Case 2}:                                                                                                                                                   \\
\textbf{Query}: According to the 2010 census, what was the population of the city in which Hosty Duo is based?                                                                                                                                                   \\
\textbf{Knowledge}: The song "Oklahoma Breakdown" was originally recorded by the Norman-based group, Hosty Duo. The population was 110,925 at the 2010 census.                                                                                                                                               \\
\textbf{Ground Truth}: 110,925                                                                                                                                            \\
\textbf{Response of LLaMA-2}: 13,250                                                                                                                                     \\
\textbf{Response of MALM}: 10,000 \\\midrule[1pt]                                                                                                                                        
\end{tabular}
    }
\end{center}
\end{table}

We also conduct an analysis of MALM's hallucination mitigation ability using selected cases in Table~\ref{tab:case}. In the first instance, the correct answer to the query is \textit{``Wheel of Time''}. However, LLaMA-2 generates \textit{``Tender Fictions''}, which does not align with the factual information. Notably, the knowledge points that \textit{``Wheel of Time''} is indeed a 2003 documentary film directed by Werner Herzog, focusing on Tibetan Buddhism. Through its generation process, MALM integrates this knowledge information and effectively mitigates hallucination, thereby generating the correct response \textit{``Wheel of Time''}.

However, MALM seems less proficient in numerical problems. In the second case, although the knowledge explains that \textit{``The population was 110,925 at the 2010 census''}, both of LLaMA-2 and MALM give incorrect answers. LLaMA-2 replies \textit{``13,250''}, and MALM replies \textit{``10,000''}. Nonetheless, the answer from MALM is closer to the ground truth, suggesting that although it does mitigate hallucination to some extent, its potential still needs to be further exploited.

\section{Discussion}
\subsection{Addressing Hallucination in Different LLMs}
The proposed MALM framework successfully mitigates hallucination problems when LLMs generate responses to questions. By using graph neural networks, we can concurrently consider the input, context and factual knowledge information and leverage the correlations between them, which are ignored in previous studies. MALM improves the performance of different LLMs with the relative improvement ranging from 0.40\% to 45.65\% in terms of ROUGE-1, from 5.07\% to 47.97\% in terms of ROUGE-2, from 0.40\% to 47.78\% in terms of ROUGE-L, from 2.33\% to 48.57\% in terms of Exact Match, and from 1.93\% to 45.00\% in terms of FEQA. Remarkably,  MALM brings a significant performance increase to LLaMA-2, which is 2.05\%, 5.10\%, 1.85\%, 3.52\% and 2.84\% respectively in terms of ROUGE-1, ROUGE-2, ROUGE-L, Exact Match and FEQA. This improvement demonstrates the potential of MALM to help different LLMs avoid hallucination problems.

\subsection{Effectiveness to Improve RAG}
We combine RAG and MALM to jointly enhance the knowledge and language comprehension ability of LLMs, which is one of the contributions of our study. We use the documents retrieved by retrievers as the knowledge information, and feed them into MALM for a better performance. The performance reaches significant improvements over the baseline models in different datasets ranging from 17.57\% to 37.49\% in terms of ROUGE-1, from 13.11\% to 23.56\% in terms of ROUGE-2, from 17.32\% to 37.31\% in terms of ROUGE-L and from 11.28\% to 27.92\% in terms of FEQA. These results demonstrate the effectiveness of MALM on RAG tasks.

\subsection{Theoretical and Practical Implications}
The theoretical contributions of this study include introducing graph neural networks into the architecture of LLMs to simultaneously consider the interdependence of three types of information, i.e., input information, context information and knowledge information, to mitigate different types of hallucination problems. We also demonstrate that MALM can effectively improve the performance of the LLMs by the external knowledge retrieved by RAG methods, and outperforms strong RAG baselines. Practically, MALM is designed as a plug-in adapter, which means it can be conveniently slotted into LLMs or removed. Moreover, it is time efficient because it can directly generate low hallucination responses, rather than a two-step approach (verification and revision) as used in most of the previous works. MALM also saves training cost, as there is no need to adjust the entire parameters of the LLMs, and only the parameters of the adapter need to be trained from scratch.

\subsection{Limitations and Future Work}
There are several limitations in our work that we intend to address in the future: (1) \textbf{Less good performance in numerical problems.} MALM seems unable to completely mitigate hallucination in numerical problems. However, the numbers can be the core in answering many users' queries, such as year, distance and size.
(2) \textbf{Reasoning Capability.} Sometimes the questions and knowledge can be puzzling, and the answers need to be reasoned. However, MALM's current architecture lacks explicit mechanisms to address the reasoning capability.
(3) \textbf{Applying Constraint.} As an plug-in adapter, MALM needs to access the hidden features of the tokens in the foundation LLMs. This limitation means that MALM can only be applied to open-source LLMs. For other types of LLMs, finding an effective method to utilize the three different types of information for hallucination mitigation remains a challenge.

\section{Conclusions}
Hallucination has become increasingly prominent in the development of LLMs, presenting a significant challenge. Hallucination can be categorized into three different types: Input-Conflicting hallucination, Context-Conflicting hallucination, and Fact-Conflicting hallucination. However, recent studies failed to simultaneously mitigate all three types of hallucination in one unified framework. To address this problem, we propose MALM as an adapter to comprehensively leverage three essential types of information related to hallucination, i.e., input information, context information and knowledge information. By constructing interactive connections between the three types of information, MALM significantly enhances the performance of a range of foundation LLMs on different evaluation metrics, demonstrating its adaptability to different LLMs. In experiments assessing generalizability for RAG, MALM proves that it can achieve the state-of-the-art performance by utilizing the external knowledge information retrieved by retrievers. In ablation study, it shows the main factors influencing MALM's performance are the structure of input connections and the knowledge information. The experiment of model layers demonstrates that the number of layers is also an important factor for MALM's performance. In type-specific mitigation analysis, we reveal that MALM can effectively mitigate all three types of hallucination and the Input-Conflicting hallucination is the most mitigated. We also conduct automated evaluation with GPT-4 and human evaluation, and the results further demonstrate the effectiveness of MALM.

\section{Acknowledgements}
Funding: This work was supported by the Natural Science Foundation of China [grant number 62376027]; and Beijing Municipal Natural Science Foundation [grant numbers IS23061 and 4222036].

\bibliographystyle{apalike}

\bibliography{cas-refs}

\begin{thebibliography}{}

\bibitem[Almazrouei et~al., 2023]{almazrouei2023falcon}
Almazrouei, E., Alobeidli, H., Alshamsi, A., Cappelli, A., Cojocaru, R.,
  Debbah, M., Goffinet, {\'E}., Hesslow, D., Launay, J., Malartic, Q., et~al.
  (2023).
\newblock The falcon series of open language models.
\newblock {\em arXiv preprint arXiv:2311.16867}.

\bibitem[Asai et~al., 2024]{asai2024selfrag}
Asai, A., Wu, Z., Wang, Y., Sil, A., and Hajishirzi, H. (2024).
\newblock Self-{RAG}: Learning to retrieve, generate, and critique through
  self-reflection.
\newblock In {\em The Twelfth International Conference on Learning
  Representations}, pages 1--30.

\bibitem[Chang et~al., 2024]{chang2024survey}
Chang, Y., Wang, X., Wang, J., Wu, Y., Yang, L., Zhu, K., Chen, H., Yi, X.,
  Wang, C., Wang, Y., et~al. (2024).
\newblock A survey on evaluation of large language models.
\newblock {\em ACM Transactions on Intelligent Systems and Technology},
  15(3):1--45.

\bibitem[Chen et~al., 2023]{chen2023purr}
Chen, A., Pasupat, P., Singh, S., Lee, H., and Guu, K. (2023).
\newblock Purr: Efficiently editing language model hallucinations by denoising
  language model corruptions.
\newblock {\em arXiv preprint arXiv:2305.14908}.

\bibitem[Chen, 2024]{chen2024entity}
Chen, J. (2024).
\newblock An entity-guided text summarization framework with relational
  heterogeneous graph neural network.
\newblock {\em Neural Computing and Applications}, 36(7):3613--3630.

\bibitem[Chen et~al., 2024a]{chen2024benchmarking}
Chen, J., Lin, H., Han, X., and Sun, L. (2024a).
\newblock Benchmarking large language models in retrieval-augmented generation.
\newblock In {\em Proceedings of the AAAI Conference on Artificial
  Intelligence}, volume~38, pages 17754--17762.

\bibitem[Chen et~al., 2024b]{chen2024unified}
Chen, X., Wang, C., Zhang, N., Xue, Y., xiaoyan yang, Shen, Y., Gu, J., and
  Chen, H. (2024b).
\newblock Unified hallucination detection for multimodal large language models.
\newblock In {\em ICLR 2024 Workshop on Reliable and Responsible Foundation
  Models}, pages 1--13.

\bibitem[Cheng et~al., 2023]{cheng2024lift}
Cheng, X., Luo, D., Chen, X., Liu, L., Zhao, D., and Yan, R. (2023).
\newblock Lift yourself up: Retrieval-augmented text generation with
  self-memory.
\newblock In {\em Advances in Neural Information Processing Systems},
  volume~36, pages 43780--43799.

\bibitem[Dhuliawala et~al., 2024]{dhuliawala2024chain}
Dhuliawala, S., Komeili, M., Xu, J., Raileanu, R., Li, X., Celikyilmaz, A., and
  Weston, J.~E. (2024).
\newblock Chain-of-verification reduces hallucination in large language models.
\newblock In {\em ICLR 2024 Workshop on Reliable and Responsible Foundation
  Models}, pages 1--19.

\bibitem[Durmus et~al., 2020]{durmus2020feqa}
Durmus, E., He, H., and Diab, M. (2020).
\newblock {FEQA}: A question answering evaluation framework for faithfulness
  assessment in abstractive summarization.
\newblock In {\em Proceedings of the 58th Annual Meeting of the Association for
  Computational Linguistics}, pages 5055--5070.

\bibitem[Edge et~al., 2024]{edge2024local}
Edge, D., Trinh, H., Cheng, N., Bradley, J., Chao, A., Mody, A., Truitt, S.,
  and Larson, J. (2024).
\newblock From local to global: A graph rag approach to query-focused
  summarization.
\newblock {\em arXiv preprint arXiv:2404.16130}.

\bibitem[Gao et~al., 2023]{gao2023rarr}
Gao, L., Dai, Z., Pasupat, P., Chen, A., Chaganty, A.~T., Fan, Y., Zhao, V.,
  Lao, N., Lee, H., Juan, D., and Guu, K. (2023).
\newblock {RARR}: Researching and revising what language models say, using
  language models.
\newblock In {\em Proceedings of the 61st Annual Meeting of the Association for
  Computational Linguistics}, volume~1, pages 16477--16508.

\bibitem[Gori et~al., 2005]{gori2005new}
Gori, M., Monfardini, G., and Scarselli, F. (2005).
\newblock A new model for learning in graph domains.
\newblock In {\em Proceedings. 2005 IEEE International Joint Conference on
  Neural Networks, 2005.}, volume~2, pages 729--734.

\bibitem[Hamilton et~al., 2017]{hamilton2017inductive}
Hamilton, W., Ying, Z., and Leskovec, J. (2017).
\newblock Inductive representation learning on large graphs.
\newblock In {\em Advances in Neural Information Processing Systems},
  volume~30, pages 1--11.

\bibitem[Hao et~al., 2024]{hao2024simplices}
Hao, Q., Wang, C., Xiao, Y., and Lin, H. (2024).
\newblock Simplices-based higher-order enhancement graph neural network for
  multi-behavior recommendation.
\newblock {\em Information Processing \& Management}, 61(5):103790.

\bibitem[He et~al., 2021]{he2021effectiveness}
He, R., Liu, L., Ye, H., Tan, Q., Ding, B., Cheng, L., Low, J., Bing, L., and
  Si, L. (2021).
\newblock On the effectiveness of adapter-based tuning for pretrained language
  model adaptation.
\newblock In {\em Proceedings of the 59th Annual Meeting of the Association for
  Computational Linguistics and the 11th International Joint Conference on
  Natural Language Processing}, volume~1, pages 2208--2222.

\bibitem[Houlsby et~al., 2019]{houlsby2019parameter}
Houlsby, N., Giurgiu, A., Jastrzebski, S., Morrone, B., De~Laroussilhe, Q.,
  Gesmundo, A., Attariyan, M., and Gelly, S. (2019).
\newblock Parameter-efficient transfer learning for {NLP}.
\newblock In {\em Proceedings of the 36th International Conference on Machine
  Learning}, volume~97, pages 2790--2799.

\bibitem[Hu et~al., 2022]{hu2022lora}
Hu, E.~J., yelong shen, Wallis, P., Allen-Zhu, Z., Li, Y., Wang, S., Wang, L.,
  and Chen, W. (2022).
\newblock Lo{RA}: Low-rank adaptation of large language models.
\newblock In {\em International Conference on Learning Representations}, pages
  1--13.

\bibitem[Huang et~al., 2024a]{huang2024opera}
Huang, Q., Dong, X., Zhang, P., Wang, B., He, C., Wang, J., Lin, D., Zhang, W.,
  and Yu, N. (2024a).
\newblock Opera: Alleviating hallucination in multi-modal large language models
  via over-trust penalty and retrospection-allocation.
\newblock In {\em Proceedings of the IEEE/CVF Conference on Computer Vision and
  Pattern Recognition}, pages 13418--13427.

\bibitem[Huang et~al., 2024b]{huang2024can}
Huang, X., Han, K., Yang, Y., Bao, D., Tao, Q., Chai, Z., and Zhu, Q. (2024b).
\newblock Can gnn be good adapter for llms?
\newblock In {\em Proceedings of the ACM on Web Conference 2024}, pages
  893--904.

\bibitem[Izacard and Grave, 2021]{izacard2021leveraging}
Izacard, G. and Grave, E. (2021).
\newblock Leveraging passage retrieval with generative models for open domain
  question answering.
\newblock In {\em Proceedings of the 16th Conference of the European Chapter of
  the Association for Computational Linguistics: Main Volume}, pages 874--880.

\bibitem[Ji et~al., 2023]{ji2023survey}
Ji, Z., Lee, N., Frieske, R., Yu, T., Su, D., Xu, Y., Ishii, E., Bang, Y.~J.,
  Madotto, A., and Fung, P. (2023).
\newblock Survey of hallucination in natural language generation.
\newblock {\em ACM Computing Surveys}, 55(12):1--38.

\bibitem[Jiang et~al., 2024]{jiang2024hallucination}
Jiang, C., Xu, H., Dong, M., Chen, J., Ye, W., Yan, M., Ye, Q., Zhang, J.,
  Huang, F., and Zhang, S. (2024).
\newblock Hallucination augmented contrastive learning for multimodal large
  language model.
\newblock In {\em Proceedings of the IEEE/CVF Conference on Computer Vision and
  Pattern Recognition}, pages 27036--27046.

\bibitem[Joshi et~al., 2017]{joshi2017triviaqa}
Joshi, M., Choi, E., Weld, D., and Zettlemoyer, L. (2017).
\newblock {T}rivia{QA}: A large scale distantly supervised challenge dataset
  for reading comprehension.
\newblock In {\em Proceedings of the 55th Annual Meeting of the Association for
  Computational Linguistics}, volume~1, pages 1601--1611.

\bibitem[Karpukhin et~al., 2020]{karpukhin2020dense}
Karpukhin, V., Oguz, B., Min, S., Lewis, P., Wu, L., Edunov, S., Chen, D., and
  Yih, W.-t. (2020).
\newblock Dense passage retrieval for open-domain question answering.
\newblock In {\em Proceedings of the 2020 Conference on Empirical Methods in
  Natural Language Processing (EMNLP)}, pages 6769--6781.

\bibitem[Kipf and Welling, 2017]{kipf2017semisupervised}
Kipf, T.~N. and Welling, M. (2017).
\newblock Semi-supervised classification with graph convolutional networks.
\newblock In {\em International Conference on Learning Representations}, pages
  1--14.

\bibitem[Kwiatkowski et~al., 2019]{kwiatkowski-etal-2019-natural}
Kwiatkowski, T., Palomaki, J., Redfield, O., Collins, M., Parikh, A., Alberti,
  C., Epstein, D., Polosukhin, I., Devlin, J., Lee, K., Toutanova, K., Jones,
  L., Kelcey, M., Chang, M.-W., Dai, A.~M., Uszkoreit, J., Le, Q., and Petrov,
  S. (2019).
\newblock Natural questions: A benchmark for question answering research.
\newblock {\em Transactions of the Association for Computational Linguistics},
  7:452--466.

\bibitem[Lee et~al., 2022]{lee2022factuality}
Lee, N., Ping, W., Xu, P., Patwary, M., Fung, P.~N., Shoeybi, M., and
  Catanzaro, B. (2022).
\newblock Factuality enhanced language models for open-ended text generation.
\newblock {\em Advances in Neural Information Processing Systems},
  35:34586--34599.

\bibitem[Leiser et~al., 2024]{leiser2024hill}
Leiser, F., Eckhardt, S., Leuthe, V., Knaeble, M., Maedche, A., Schwabe, G.,
  and Sunyaev, A. (2024).
\newblock Hill: A hallucination identifier for large language models.
\newblock In {\em Proceedings of the CHI Conference on Human Factors in
  Computing Systems}, pages 1--13.

\bibitem[Li et~al., 2024]{li2024homogeneous}
Li, D., Gao, Y., Wang, Z., Qiu, H., Liu, P., Xiong, Z., and Zhang, Z. (2024).
\newblock Homogeneous graph neural networks for third-party library
  recommendation.
\newblock {\em Information Processing \& Management}, 61(6):103831.

\bibitem[Li et~al., 2023]{li2023halueval}
Li, J., Cheng, X., Zhao, W.~X., Nie, J., and Wen, J. (2023).
\newblock Halueval: A large-scale hallucination evaluation benchmark for large
  language models.
\newblock In {\em Proceedings of the 2023 Conference on Empirical Methods in
  Natural Language Processing}, pages 6449--6464.

\bibitem[Lin, 2004]{lin2004rouge}
Lin, C. (2004).
\newblock Rouge: A package for automatic evaluation of summaries.
\newblock In {\em Text summarization branches out}, pages 74--81.

\bibitem[Lin et~al., 2022]{lin2022truthfulqa}
Lin, S., Hilton, J., and Evans, O. (2022).
\newblock Truthfulqa: Measuring how models mimic human falsehoods.
\newblock In {\em Proceedings of the 60th Annual Meeting of the Association for
  Computational Linguistics (Volume 1: Long Papers)}, pages 3214--3252.

\bibitem[Liu et~al., 2025]{liu2025evopath}
Liu, S., Cheng, H., Wang, Y., He, Y., Fan, C., and Liu, Z. (2025).
\newblock Evopath: Evolutionary meta-path discovery with large language models
  for complex heterogeneous information networks.
\newblock {\em Information Processing \& Management}, 62(1):103920.

\bibitem[Maazallahi et~al., 2025]{maazallahi2025advancing}
Maazallahi, A., Asadpour, M., and Bazmi, P. (2025).
\newblock Advancing emotion recognition in social media: A novel integration of
  heterogeneous neural networks with fine-tuned language models.
\newblock {\em Information Processing \& Management}, 62(2):103974.

\bibitem[Manakul et~al., 2023]{manakul2023selfcheckgpt}
Manakul, P., Liusie, A., and Gales, M. (2023).
\newblock {S}elf{C}heck{GPT}: Zero-resource black-box hallucination detection
  for generative large language models.
\newblock In {\em Proceedings of the 2023 Conference on Empirical Methods in
  Natural Language Processing}, pages 9004--9017.

\bibitem[McDonald et~al., 2024]{mcdonald2024reducing}
McDonald, D., Papadopoulos, R., and Benningfield, L. (2024).
\newblock Reducing llm hallucination using knowledge distillation: A case study
  with mistral large and mmlu benchmark.
\newblock {\em TechRxiv}.

\bibitem[Meng et~al., 2022]{meng2022gnn}
Meng, Y., Zong, S., Li, X., Sun, X., Zhang, T., Wu, F., and Li, J. (2022).
\newblock {GNN}-{LM}: Language modeling based on global contexts via {GNN}.
\newblock In {\em ICLR 2022 Workshop on Deep Learning on Graphs for Natural
  Language Processing}, pages 1--13.

\bibitem[M{\"u}ndler et~al., 2024]{mundler2024self}
M{\"u}ndler, N., He, J., Jenko, S., and Vechev, M. (2024).
\newblock Self-contradictory hallucinations of large language models:
  Evaluation, detection and mitigation.
\newblock In {\em The Twelfth International Conference on Learning
  Representations}, pages 1--30.

\bibitem[OpenAI, 2023]{openai2023gpt}
OpenAI (2023).
\newblock Gpt-4 technical report.
\newblock {\em arXiv preprint arXiv:2303.08774}.

\bibitem[Papineni et~al., 2002]{papineni2002bleu}
Papineni, K., Roukos, S., Ward, T., and Zhu, W. (2002).
\newblock {BLEU}: a {M}ethod for {A}utomatic {E}valuation of {M}achine
  {T}ranslation.
\newblock In {\em Proceedings of the 40th annual meeting of the Association for
  Computational Linguistics}, pages 311--318.

\bibitem[Pfeiffer et~al., 2023]{pfeiffer2023mmt5}
Pfeiffer, J., Piccinno, F., Nicosia, M., Wang, X., Reid, M., and Ruder, S.
  (2023).
\newblock mm{T}5: {M}odular {M}ultilingual {P}re-{T}raining {S}olves {S}ource
  {L}anguage {H}allucinations.
\newblock In {\em Findings of the Association for Computational Linguistics:
  EMNLP 2023}, pages 1978--2008.

\bibitem[Radford et~al., 2019]{radford2019language}
Radford, A., Wu, J., Child, R., Luan, D., Amodei, D., and Sutskever, I. (2019).
\newblock Language models are unsupervised multitask learners.
\newblock {\em OpenAI blog}, 1(8):9.

\bibitem[Rajpurkar et~al., 2016]{rajpurkar2016squad}
Rajpurkar, P., Zhang, J., Lopyrev, K., and Liang, P. (2016).
\newblock {SQ}u{AD}: 100,000+ {Q}uestions for {M}achine {C}omprehension of
  {T}ext.
\newblock {\em arXiv preprint arXiv:1606.05250}.

\bibitem[Ram et~al., 2023]{ram-etal-2023-context}
Ram, O., Levine, Y., Dalmedigos, I., Muhlgay, D., Shashua, A., Leyton-Brown,
  K., and Shoham, Y. (2023).
\newblock In-{C}ontext {R}etrieval-{A}ugmented {L}anguage {M}odels.
\newblock {\em Transactions of the Association for Computational Linguistics},
  11:1316--1331.

\bibitem[Ram et~al., 2022]{ram2022learning}
Ram, O., Shachaf, G., Levy, O., Berant, J., and Globerson, A. (2022).
\newblock Learning to {R}etrieve {P}assages without {S}upervision.
\newblock In {\em Proceedings of the 2022 Conference of the North American
  Chapter of the Association for Computational Linguistics: Human Language
  Technologies}, pages 2687--2700.

\bibitem[Ramrakhiyani et~al., 2025]{ramrakhiyani2025gauging}
Ramrakhiyani, N., Varma, V., Palshikar, G.~K., and Pawar, S. (2025).
\newblock Gauging, enriching and applying geography knowledge in pre-trained
  language models.
\newblock {\em Information Processing \& Management}, 62(1):103892.

\bibitem[Robertson and Zaragoza, 2009]{robertson2009probabilistic}
Robertson, S. and Zaragoza, H. (2009).
\newblock The {P}robabilistic {R}elevance {F}ramework: {BM}25 and {B}eyond.
\newblock {\em Information Retrieval}, 3(4):333--389.

\bibitem[Sarthi et~al., 2024]{sarthi2024raptor}
Sarthi, P., Abdullah, S., Tuli, A., Khanna, S., Goldie, A., and Manning, C.~D.
  (2024).
\newblock {RAPTOR}: {R}ecursive {A}bstractive {P}rocessing for
  {T}ree-{O}rganized {R}etrieval.
\newblock In {\em The Twelfth International Conference on Learning
  Representations}, pages 1--22.

\bibitem[Scao et~al., 2022]{workshop2022bloom}
Scao, T.~L., Fan, A., Akiki, C., Pavlick, E., Ili{\'c}, S., Hesslow, D.,
  Castagn{\'e}, R., Luccioni, A.~S., Yvon, F., Gall{\'e}, M., Tow, J., Rush,
  A.~M., Biderman, S., Webson, A., Ammanamanchi, P.~S., Wang, T., Sagot, B.,
  Muennighoff, N., del Moral, A.~V., Ruwase, O., Bawden, R., Bekman, S.,
  Mcmillan-Major, A., Beltagy, I., Nguyen, H., Saulnier, L., Tan, S.,
  Ortiz~Suarez, P., Sanh, V., Lauren{\c c}on, H., Jernite, Y., Launay, J.,
  Mitchell, M., Raffel, C., Gokaslan, A., Simhi, A., Soroa, A., Aji, A.~F.,
  Alfassy, A., Rogers, A., Nitzav, A.~K., Xu, C., Mou, C., Emezue, C., Klamm,
  C., Leong, C., van Strien, D., Adelani, D.~I., Radev, D., Ponferrada, E.~G.,
  Levkovizh, E., Kim, E., Natan, E.~B., de~Toni, F., Dupont, G., Kruszewski,
  G., Pistilli, G., Elsahar, H., Benyamina, H., Tran, H., Yu, I., Abdulmumin,
  I., Johnson, I., Gonzalez-Dios, I., de~la Rosa, J., Chim, J., Dodge, J., Zhu,
  J., Chang, J., Frohberg, J., Tobing, J., Bhattacharjee, J., Almubarak, K.,
  Chen, K., Lo, K., von Werra, L., Weber, L., Phan, L., Allal, L.~B., Tanguy,
  L., Dey, M., Mu{\~n}oz, M.~R., Masoud, M., Grandury, M., {\v S}a{\v s}ko, M.,
  Huang, M., Coavoux, M., Singh, M., Jiang, M. T.-J., Vu, M.~C., Jauhar, M.~A.,
  Ghaleb, M., Subramani, N., Kassner, N., Khamis, N., Nguyen, O., Espejel, O.,
  de~Gibert, O., Villegas, P., Henderson, P., Colombo, P., Amuok, P., Lhoest,
  Q., Harliman, R., Bommasani, R., L{\'o}pez, R.~L., Ribeiro, R., Osei, S.,
  Pyysalo, S., Nagel, S., Bose, S., Muhammad, S.~H., Sharma, S., Longpre, S.,
  Nikpoor, S., Silberberg, S., Pai, S., Zink, S., Torrent, T.~T., Schick, T.,
  Thrush, T., Danchev, V., Nikoulina, V., Laippala, V., Lepercq, V., Prabhu,
  V., Alyafeai, Z., Talat, Z., Raja, A., Heinzerling, B., Si, C., Salesky, E.,
  Mielke, S.~J., Lee, W.~Y., Sharma, A., Santilli, A., Chaffin, A., Stiegler,
  A., Datta, D., Szczechla, E., Chhablani, G., Wang, H., Pandey, H., Strobelt,
  H., Fries, J.~A., Rozen, J., Gao, L., Sutawika, L., Bari, M.~S., Al-Shaibani,
  M.~S., Manica, M., Nayak, N., Teehan, R., Albanie, S., Shen, S., Ben-David,
  S., Bach, S.~H., Kim, T., Bers, T., Fevry, T., Neeraj, T., Thakker, U.,
  Raunak, V., Tang, X., Yong, Z.-X., Sun, Z., Brody, S., Uri, Y., Tojarieh, H.,
  Roberts, A., Chung, H.~W., Tae, J., Phang, J., Press, O., Li, C., Narayanan,
  D., Bourfoune, H., Casper, J., Rasley, J., Ryabinin, M., Mishra, M., Zhang,
  M., Shoeybi, M., Peyrounette, M., Patry, N., Tazi, N., Sanseviero, O., von
  Platen, P., Cornette, P., Lavall{\'e}e, P.~F., Lacroix, R., Rajbhandari, S.,
  Gandhi, S., Smith, S., Requena, S., Patil, S., Dettmers, T., Baruwa, A.,
  Singh, A., Cheveleva, A., Ligozat, A.-L., Subramonian, A., N{\'e}v{\'e}ol,
  A., Lovering, C., Garrette, D., Tunuguntla, D., Reiter, E., Taktasheva, E.,
  Voloshina, E., Bogdanov, E., Winata, G.~I., Schoelkopf, H., Kalo, J.-C.,
  Novikova, J., Forde, J.~Z., Clive, J., Kasai, J., Kawamura, K., Hazan, L.,
  Carpuat, M., Clinciu, M., Kim, N., Cheng, N., Serikov, O., Antverg, O.,
  van~der Wal, O., Zhang, R., Zhang, R., Gehrmann, S., Pais, S., Shavrina, T.,
  Scialom, T., Yun, T., Limisiewicz, T., Rieser, V., Protasov, V., Mikhailov,
  V., Pruksachatkun, Y., Belinkov, Y., Bamberger, Z., Kasner, Z., Rueda, A.,
  Pestana, A., Feizpour, A., Khan, A., Faranak, A., Santos, A., Hevia, A.,
  Unldreaj, A., Aghagol, A., Abdollahi, A., Tammour, A., Hajihosseini, A.,
  Behroozi, B., Ajibade, B., Saxena, B., Ferrandis, C.~M., Contractor, D.,
  Lansky, D., David, D., Kiela, D., Nguyen, D.~A., Tan, E., Baylor, E., Ozoani,
  E., Mirza, F., Ononiwu, F., Rezanejad, H., Jones, H., Bhattacharya, I.,
  Solaiman, I., Sedenko, I., Nejadgholi, I., Passmore, J., Seltzer, J., Sanz,
  J.~B., Dutra, L., Samagaio, M., Elbadri, M., Mieskes, M., Gerchick, M.,
  Akinlolu, M., Mckenna, M., Qiu, M., Ghauri, M., Burynok, M., Abrar, N.,
  Rajani, N., Elkott, N., Fahmy, N., Samuel, O., An, R., Kromann, R., Hao, R.,
  Alizadeh, S., Shubber, S., Wang, S., Roy, S., Viguier, S., Le, T., Oyebade,
  T., Le, T., Yang, Y., Nguyen, Z., Kashyap, A.~R., Palasciano, A., Callahan,
  A., Shukla, A., Miranda-Escalada, A., Singh, A., Beilharz, B., Wang, B.,
  Brito, C., Zhou, C., Jain, C., Xu, C., Fourrier, C., Peri{\~n}{\'a}n, D.~L.,
  Molano, D., Yu, D., Manjavacas, E., Barth, F., Fuhrimann, F., Altay, G.,
  Bayrak, G., Burns, G., Vrabec, H.~U., Bello, I., Dash, I., Kang, J., Giorgi,
  J., Golde, J., Posada, J.~D., Sivaraman, K.~R., Bulchandani, L., Liu, L.,
  Shinzato, L., de~Bykhovetz, M.~H., Takeuchi, M., P{\`a}mies, M., Castillo,
  M.~A., Nezhurina, M., S{\"a}nger, M., Samwald, M., Cullan, M., Weinberg, M.,
  de~Wolf, M., Mihaljcic, M., Liu, M., Freidank, M., Kang, M., Seelam, N.,
  Dahlberg, N., Broad, N.~M., Muellner, N., Fung, P., Haller, P.,
  Chandrasekhar, R., Eisenberg, R., Martin, R., Canalli, R., Su, R., Su, R.,
  Cahyawijaya, S., Garda, S., Deshmukh, S.~S., Mishra, S., Kiblawi, S., Ott,
  S., Sang-Aroonsiri, S., Kumar, S., Schweter, S., Bharati, S., Laud, T.,
  Gigant, T., Kainuma, T., Kusa, W., Labrak, Y., Bajaj, Y.~S., Venkatraman, Y.,
  Xu, Y., Xu, Y., Xu, Y., Tan, Z., Xie, Z., Ye, Z., Bras, M., Belkada, Y., and
  Wolf, T. (2022).
\newblock {BLOOM}: {A} 176{B}-{P}arameter {O}pen-{A}ccess {M}ultilingual
  {L}anguage {M}odel.
\newblock {\em arXiv preprint arXiv:2211.05100}.

\bibitem[Shi et~al., 2024]{shi-etal-2024-replug}
Shi, W., Min, S., Yasunaga, M., Seo, M., James, R., Lewis, M., Zettlemoyer, L.,
  and Yih, W.-t. (2024).
\newblock {REPLUG}: {R}etrieval-{A}ugmented {B}lack-{B}ox {L}anguage {M}odels.
\newblock In {\em Proceedings of the 2024 Conference of the North American
  Chapter of the Association for Computational Linguistics: Human Language
  Technologies}, volume~1, pages 8371--8384.

\bibitem[Sun et~al., 2023]{sun2023contrastive}
Sun, W., Shi, Z., Gao, S., Ren, P., de~Rijke, M., and Ren, Z. (2023).
\newblock Contrastive learning reduces hallucination in conversations.
\newblock In {\em Proceedings of the AAAI Conference on Artificial
  Intelligence}, volume~37, pages 13618--13626.

\bibitem[Touvron et~al., 2023a]{touvron2023llama}
Touvron, H., Lavril, T., Izacard, G., Martinet, X., Lachaux, M.-A., Lacroix,
  T., Rozière, B., Goyal, N., Hambro, E., Azhar, F., Rodriguez, A., Joulin,
  A., Grave, E., and Lample, G. (2023a).
\newblock Llama: Open and efficient foundation language models.
\newblock {\em arXiv preprint arXiv:2302.13971}.

\bibitem[Touvron et~al., 2023b]{touvron2023llama2}
Touvron, H., Martin, L., Stone, K., Albert, P., Almahairi, A., Babaei, Y.,
  Bashlykov, N., Batra, S., Bhargava, P., Bhosale, S., Bikel, D., Blecher, L.,
  Ferrer, C.~C., Chen, M., Cucurull, G., Esiobu, D., Fernandes, J., Fu, J., Fu,
  W., Fuller, B., Gao, C., Goswami, V., Goyal, N., Hartshorn, A., Hosseini, S.,
  Hou, R., Inan, H., Kardas, M., Kerkez, V., Khabsa, M., Kloumann, I., Korenev,
  A., Koura, P.~S., Lachaux, M.-A., Lavril, T., Lee, J., Liskovich, D., Lu, Y.,
  Mao, Y., Martinet, X., Mihaylov, T., Mishra, P., Molybog, I., Nie, Y.,
  Poulton, A., Reizenstein, J., Rungta, R., Saladi, K., Schelten, A., Silva,
  R., Smith, E.~M., Subramanian, R., Tan, X.~E., Tang, B., Taylor, R.,
  Williams, A., Kuan, J.~X., Xu, P., Yan, Z., Zarov, I., Zhang, Y., Fan, A.,
  Kambadur, M., Narang, S., Rodriguez, A., Stojnic, R., Edunov, S., and
  Scialom, T. (2023b).
\newblock Llama 2: Open foundation and fine-tuned chat models.
\newblock {\em arXiv preprint arXiv:2307.09288}.

\bibitem[Varshney et~al., 2023]{varshney2023stitch}
Varshney, N., Yao, W., Zhang, H., Chen, J., and Yu, D. (2023).
\newblock A stitch in time saves nine: Detecting and mitigating hallucinations
  of llms by validating low-confidence generation.
\newblock {\em arXiv preprint arXiv:2307.03987}.

\bibitem[Veličković et~al., 2018]{velickovic2018graph}
Veličković, P., Cucurull, G., Casanova, A., Romero, A., Liò, P., and Bengio,
  Y. (2018).
\newblock Graph attention networks.
\newblock In {\em International Conference on Learning Representations}, pages
  1--12.

\bibitem[Wang et~al., 2024]{wang2024searching}
Wang, X., Wang, Z., Gao, X., Zhang, F., Wu, Y., Xu, Z., Shi, T., Wang, Z., Li,
  S., Qian, Q., et~al. (2024).
\newblock Searching for best practices in retrieval-augmented generation.
\newblock In {\em Proceedings of the 2024 Conference on Empirical Methods in
  Natural Language Processing}, pages 17716--17736.

\bibitem[Wang et~al., 2025]{wang2024efficient}
Wang, Y., Hu, X., Gan, Q., Huang, X., Qiu, X., and Wipf, D. (2025).
\newblock Efficient {L}ink {P}rediction via {GNN} {L}ayers {I}nduced by
  {N}egative {S}ampling.
\newblock {\em IEEE Transactions on Knowledge and Data Engineering},
  37(1):253--264.

\bibitem[Wu et~al., 2025]{wu2025asymmetric}
Wu, Z., Chen, J., Al-Sabri, R., Oloulade, B.~M., and Gao, J. (2025).
\newblock Asymmetric augmented paradigm-based graph neural architecture search.
\newblock {\em Information Processing \& Management}, 62(1):103897.

\bibitem[Wu et~al., 2020]{wu2020comprehensive}
Wu, Z., Pan, S., Chen, F., Long, G., Zhang, C., and Philip, S.~Y. (2020).
\newblock A comprehensive survey on graph neural networks.
\newblock {\em IEEE transactions on neural networks and learning systems},
  32(1):4--24.

\bibitem[Xing and Tsang, 2022]{xing2022darer}
Xing, B. and Tsang, I. (2022).
\newblock {DARER}: {D}ual-task {T}emporal {R}elational {R}ecurrent {R}easoning
  {N}etwork for {J}oint {D}ialog {S}entiment {C}lassification and {A}ct
  {R}ecognition.
\newblock In {\em Findings of the Association for Computational Linguistics:
  ACL 2022}, pages 3611--3621.

\bibitem[Xu et~al., 2023]{xu2023recomp}
Xu, F., Shi, W., and Choi, E. (2023).
\newblock Recomp: Improving retrieval-augmented lms with compression and
  selective augmentation.
\newblock {\em arXiv preprint arXiv:2310.04408}.

\bibitem[Yan et~al., 2024]{yan2024corrective}
Yan, S.-Q., Gu, J.-C., Zhu, Y., and Ling, Z.-H. (2024).
\newblock Corrective retrieval augmented generation.
\newblock {\em arXiv preprint arXiv:2401.15884}.

\bibitem[Yang et~al., 2024]{yang2024qwen2technicalreport}
Yang, A., Yang, B., Hui, B., Zheng, B., Yu, B., Zhou, C., Li, C., Li, C., Liu,
  D., Huang, F., Dong, G., Wei, H., Lin, H., Tang, J., Wang, J., Yang, J., Tu,
  J., Zhang, J., Ma, J., Yang, J., Xu, J., Zhou, J., Bai, J., He, J., Lin, J.,
  Dang, K., Lu, K., Chen, K., Yang, K., Li, M., Xue, M., Ni, N., Zhang, P.,
  Wang, P., Peng, R., Men, R., Gao, R., Lin, R., Wang, S., Bai, S., Tan, S.,
  Zhu, T., Li, T., Liu, T., Ge, W., Deng, X., Zhou, X., Ren, X., Zhang, X.,
  Wei, X., Ren, X., Liu, X., Fan, Y., Yao, Y., Zhang, Y., Wan, Y., Chu, Y.,
  Liu, Y., Cui, Z., Zhang, Z., Guo, Z., and Fan, Z. (2024).
\newblock Qwen2 technical report.
\newblock {\em arXiv preprint arXiv:2407.10671}.

\bibitem[Yang et~al., 2025]{qwen2025qwen25technicalreport}
Yang, A., Yang, B., Zhang, B., Hui, B., Zheng, B., Yu, B., Li, C., Liu, D.,
  Huang, F., Wei, H., Lin, H., Yang, J., Tu, J., Zhang, J., Yang, J., Yang, J.,
  Zhou, J., Lin, J., Dang, K., Lu, K., Bao, K., Yang, K., Yu, L., Li, M., Xue,
  M., Zhang, P., Zhu, Q., Men, R., Lin, R., Li, T., Tang, T., Xia, T., Ren, X.,
  Ren, X., Fan, Y., Su, Y., Zhang, Y., Wan, Y., Liu, Y., Cui, Z., Zhang, Z.,
  and Qiu, Z. (2025).
\newblock Qwen2.5 technical report.
\newblock {\em arXiv preprint arXiv:2412.15115}.

\bibitem[Yin and Zhong, 2024]{yin2024textgt}
Yin, S. and Zhong, G. (2024).
\newblock Text{GT}: {A} {D}ouble-{V}iew {G}raph {T}ransformer on {T}ext for
  {A}spect-{B}ased {S}entiment {A}nalysis.
\newblock In {\em Proceedings of the AAAI Conference on Artificial
  Intelligence}, volume~38, pages 19404--19412.

\bibitem[Zhang et~al., 2023a]{zhang2023m3gat}
Zhang, Y., Jia, A., Wang, B., Zhang, P., Zhao, D., Li, P., Hou, Y., Jin, X.,
  Song, D., and Qin, J. (2023a).
\newblock M3{GAT}: {A} {M}ulti-{M}odal {M}ulti-{T}ask {I}nteractive {G}raph
  {A}ttention {N}etwork for {C}onversational {S}entiment {A}nalysis and
  {E}motion {R}ecognition.
\newblock {\em ACM Transactions on Information Systems}, 42(1):1--32.

\bibitem[Zhang et~al., 2023b]{zhang2023siren}
Zhang, Y., Li, Y., Cui, L., Cai, D., Liu, L., Fu, T., Huang, X., Zhao, E.,
  Zhang, Y., Chen, Y., Wang, L., Luu, A.~T., Bi, W., Shi, F., and Shi, S.
  (2023b).
\newblock Siren's song in the ai ocean: A survey on hallucination in large
  language models.
\newblock {\em arXiv preprint arXiv:2309.01219}.

\bibitem[Zheng et~al., 2023]{zheng2023judging}
Zheng, L., Chiang, W.-L., Sheng, Y., Zhuang, S., Wu, Z., Zhuang, Y., Lin, Z.,
  Li, Z., Li, D., Xing, E., Zhang, H., Gonzalez, J.~E., and Stoica, I. (2023).
\newblock Judging {LLM}-as-a-{J}udge with {MT}-{B}ench and {C}hatbot {A}rena.
\newblock In {\em Advances in Neural Information Processing Systems},
  volume~36, pages 46595--46623.

\bibitem[Zhou et~al., 2023]{zhou2023least}
Zhou, D., Sch{\"a}rli, N., Hou, L., Wei, J., Scales, N., Wang, X., Schuurmans,
  D., Cui, C., Bousquet, O., Le, Q.~V., and Chi, E.~H. (2023).
\newblock Least-to-{M}ost {P}rompting {E}nables {C}omplex {R}easoning in
  {L}arge {L}anguage {M}odels.
\newblock In {\em The Eleventh International Conference on Learning
  Representations}, pages 1--61.

\bibitem[Zhou et~al., 2020]{zhou2020graph}
Zhou, J., Cui, G., Hu, S., Zhang, Z., Yang, C., Liu, Z., Wang, L., Li, C., and
  Sun, M. (2020).
\newblock Graph neural networks: {A} review of methods and applications.
\newblock {\em AI Open}, 1:57--81.

\end{thebibliography}





\end{document}